\documentclass[runningheads]{llncs}

\usepackage{eccv}

\usepackage{eccvabbrv}

\usepackage{graphicx}
\usepackage{booktabs}
\usepackage{multirow}
\usepackage[table]{xcolor}
\usepackage{makecell}

\newcommand{\best}[1]{\textbf{#1}}
\newcommand{\second}[1]{\underline{#1}}
\newcommand{\ours}{\textbf{AnchorPrune}}
\newcommand{\ourrow}{\rowcolor{gray!10}}
\usepackage{algorithm}
\usepackage{algpseudocode}

\usepackage[accsupp]{axessibility}  % Improves PDF readability for those with disabilities.

\usepackage[colorlinks=true,
            linkcolor=blue,
            citecolor=blue,
            urlcolor=magenta]{hyperref}

\usepackage{orcidlink}

\usepackage{booktabs}
\usepackage{graphicx}
\usepackage[table]{xcolor}

\makeatletter
\def\@fnsymbol#1{\ensuremath{\ifcase#1\or \dagger\or \ddagger\or \mathsection\or \mathparagraph\or \|\or **\or \dagger\dagger \or \ddagger\ddagger \else\@ctrerr\fi}}
\makeatother

\begin{document}

\title{AnchorPrune: Relevance-Anchored Contextual Expansion for Visual Token Pruning} 

\titlerunning{AnchorPrune}

\author{Kyuan Oh\orcidlink{0009-0001-8047-6396} \and
Bumsoo Kim\thanks{: corresponding author}\orcidlink{0009-0003-6519-3253}}

\authorrunning{K.~Oh and B.~Kim}

\institute{Chung-Ang University, Seoul, Korea \\
\email{\{oka04108,bumsoo\}@cau.ac.kr}}

\maketitle

\begin{abstract}
Large vision-language models incur substantial inference costs because high-resolution inputs introduce thousands of visual tokens, many of which are redundant for a given query. Existing pruning methods often combine query relevance and token diversity, yet these objectives can conflict under aggressive compression: relevance-driven selection may overconcentrate the budget on correlated local evidence, while diversity-driven selection may suppress indispensable tokens or retain distinct but uninformative regions. We introduce \textbf{AnchorPrune}, a training-free framework that first constructs a protected relevance anchor and then expands it with complementary visual context. AnchorPrune adaptively determines the anchor size from the novelty profile of relevance-ranked tokens, preserving a compact set of query-critical evidence, and allocates the remaining budget through importance-weighted novelty to recover informative, non-redundant context relative to the anchor. This ordered design prevents contextual expansion from displacing indispensable query cues while improving overall visual coverage. AnchorPrune is lightweight, architecture-aware, and requires neither retraining nor model modification. Across image and video vision-language models and benchmarks, it consistently improves the accuracy--efficiency trade-off over training-free baselines, particularly under severe compression. On LLaVA-NeXT-7B, AnchorPrune preserves 97.6\% of full-token performance using only 160 of 2,880 visual tokens. These results establish relevance-anchored contextual expansion as an effective principle for efficient multimodal inference. Code is available at \url{https://github.com/MULTI-cau/AnchorPrune}.
\end{abstract}

\section{Introduction}
\label{sec:introduction}

Vision-language models (VLMs) achieve strong performance across image and video understanding tasks, but their long visual-token sequences incur substantial inference costs~\cite{llava,llavanext,qwen,llava_video}. High-resolution, multi-crop, and multi-frame inputs can introduce thousands of visual tokens, many of which are redundant for a given query yet must still be processed by the language model. This overhead is especially pronounced during prefilling, where visual tokens dominate sequence length, memory consumption, and attention computation. Visual token pruning therefore provides a direct route to efficient multimodal inference by retaining only the evidence needed for accurate prediction~\cite{fastv,pyramiddrop,sparsevlm,prumerge,visionzip,divprune,cdpruner}.

\begin{figure*}[!t]
\centering
\includegraphics[width=\textwidth]{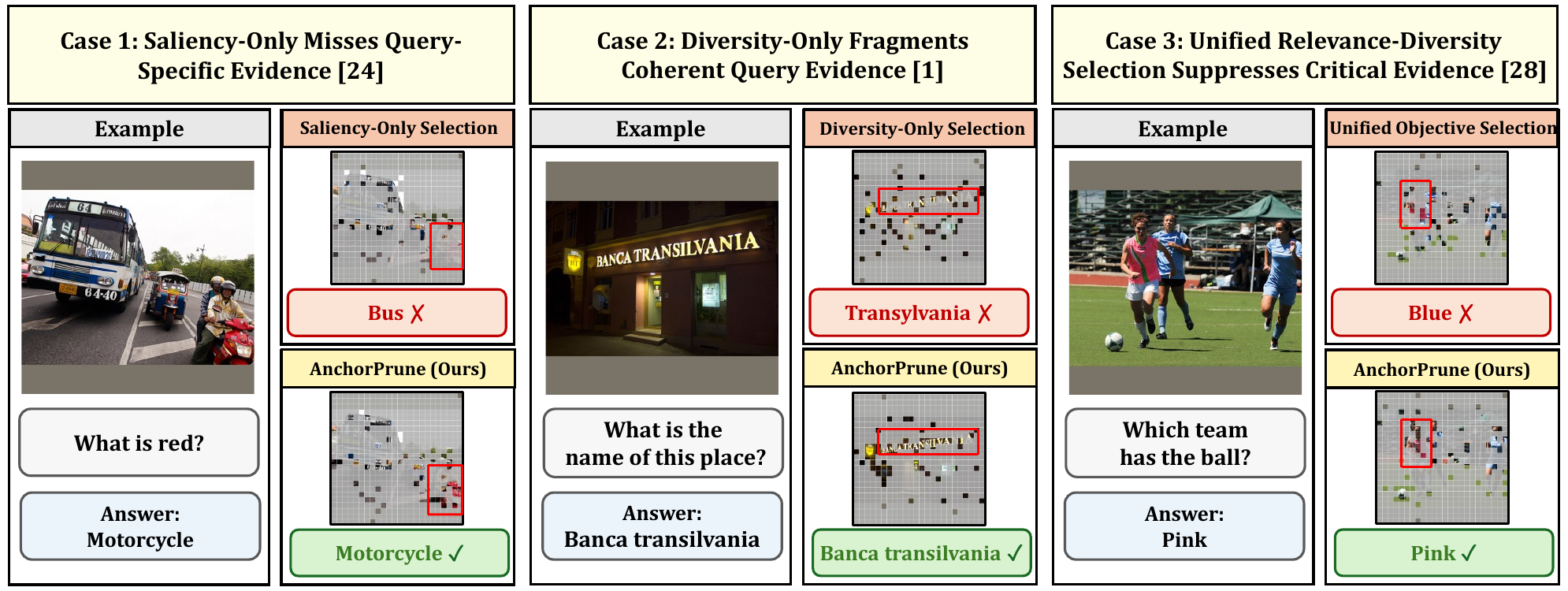}
\caption{\textbf{Qualitative comparison on LLaVA-1.5-7B with 64 retained visual tokens.}
VisionZip~\cite{visionzip} uses query-agnostic saliency, DivPrune~\cite{divprune} emphasizes query-agnostic diversity, and CDPruner~\cite{cdpruner} combines relevance and diversity within a unified objective. Under severe compression, these strategies may miss, fragment, or suppress query-critical evidence. AnchorPrune instead protects a relevance anchor before expanding it with informative, non-redundant context, yielding correct predictions across diverse tasks.}
\label{fig:teaser_qualitative}
\end{figure*}

The central challenge is not merely estimating token importance, but determining which evidence must be protected before the remaining budget is allocated. As illustrated in Fig.~\ref{fig:teaser_qualitative}, saliency-based methods may retain redundant high-importance regions while overlooking query-specific cues~\cite{fastv,pyramiddrop,sparsevlm,prumerge,visionzip}. Diversity-based methods distribute selections across the visual feature space, but can fragment locally coherent evidence or preserve visually distinct yet query-irrelevant regions~\cite{divprune}. Query-aware methods improve instruction alignment, but relevance-driven selection may still overconcentrate the budget on correlated local evidence. Recent approaches address this limitation by combining relevance and diversity within a unified objective~\cite{cdpruner}. Under aggressive compression, however, such coupling provides no explicit protection for indispensable evidence: diversity can displace query-critical tokens, while weak or diffuse relevance guidance cannot be compensated by a separate contextual expansion stage.

We argue that query-critical evidence and complementary context play asymmetric roles and should therefore be selected in a fixed order. Evidence directly required by the query, such as a small object, scene text, fine-grained attribute, or spatial relation, is largely non-substitutable: once removed, it cannot be recovered by subsequent selection. Contextual evidence, by contrast, is more substitutable, as multiple informative subsets may provide comparable support for grounding, disambiguation, or holistic reasoning. Effective pruning should therefore first establish a protected relevance anchor and only then expand it with complementary context. This expansion should prioritize tokens that are both informative and non-redundant with respect to the retained anchor, rather than rewarding visual distinctiveness alone.

Based on this principle, we introduce \textbf{AnchorPrune}, a training-free framework for relevance-anchored contextual expansion. AnchorPrune first ranks visual tokens using architecture-appropriate query-conditioned priority scores and constructs a protected relevance anchor. Rather than assigning a fixed fraction of the token budget to relevance, it determines the anchor size adaptively from the novelty profile of relevance-ranked tokens, accommodating both concentrated and distributed query evidence. The remaining budget is then allocated through importance-weighted novelty, expanding the retained set with tokens that are globally informative and complementary to the current selection. By separating anchor construction from contextual expansion, AnchorPrune protects indispensable query evidence while allowing broader context to compensate when relevance guidance is incomplete or spatially diffuse.

AnchorPrune requires neither retraining nor architectural modification and applies to both CLIP-aligned and non-CLIP VLMs using architecture-specific representation spaces. Across image and video benchmarks, it consistently improves the accuracy--efficiency trade-off over representative training-free pruning methods, with the largest gains under aggressive compression. On LLaVA-NeXT-7B, AnchorPrune preserves 97.6\% of full-token performance while retaining only 160 of 2,880 visual tokens. These results establish relevance-anchored contextual expansion as an effective principle for efficient multimodal inference.

Our contributions are threefold:
\begin{itemize}
\item We identify an ordering limitation in existing visual-token pruning methods: jointly optimizing relevance and diversity can both displace non-substitutable query evidence and leave weak or diffuse relevance guidance without a separate mechanism for contextual compensation.
\item We introduce \textbf{AnchorPrune}, a training-free framework that constructs an adaptive protected relevance anchor and expands it through importance-weighted novelty to retain informative, non-redundant context.
\item We validate AnchorPrune across CLIP-aligned and non-CLIP image and video VLMs, demonstrating consistent performance preservation and strong accuracy--efficiency trade-offs under severe visual-token compression.
\end{itemize}

\section{Preliminaries}
\label{sec:preliminaries}

\subsection{Visual Token Pruning in VLMs}
\label{sec:prelim_token_pruning}

Given a visual input $I$ and a text query $q$, a vision encoder produces a sequence of visual tokens
\begin{equation}
X=\{x_i\}_{i=1}^{N}, \qquad x_i\in\mathbb{R}^{d}.
\end{equation}
Visual token pruning selects an index set $S \subseteq \{1,\ldots,N\}$ with $|S|=K\ll N$ and passes the reduced sequence $X_S = \{x_i : i \in S\}$ to the language model. The objective is to reduce inference cost while approximately preserving the conditional output distribution:
\begin{equation}
p_{\theta}(y\mid X,q)
\approx
p_{\theta}(y\mid X_S,q).
\end{equation}
Because direct subset optimization is combinatorial, existing methods rely on surrogate criteria such as token importance, query relevance, and feature diversity.

\subsection{Relevance, Diversity, and Contextual Expansion}
\label{sec:prelim_relevance_coverage}

Relevance-driven selection prioritizes instruction-aligned evidence, but may concentrate the retained budget on correlated local tokens. Diversity-driven selection instead favors candidates that contribute information not already represented by the retained set. Given visual features $h_i$, the novelty of candidate token $i$ relative to a selected set $S$ can be measured as
\begin{equation}
\Delta(i;S)
=
\min_{j\in S}
\left(
1-
\frac{h_i^{\top}h_j}
{\lVert h_i\rVert_2\lVert h_j\rVert_2}
\right).
\end{equation}

Recent query-aware methods combine relevance and diversity within a unified subset-selection objective~\cite{cdpruner}. For example, let $\widetilde{r}\in\mathbb{R}^{N}$ denote normalized token-relevance scores. A visual-similarity kernel $L$ can be conditioned on relevance as
\begin{equation}
\widetilde{L}
=
\operatorname{diag}(\widetilde{r})
L
\operatorname{diag}(\widetilde{r}),
\qquad
S^{\star}
=
\arg\max_{|S|=K}
\log\det(\widetilde{L}_S).
\end{equation}
This formulation jointly promotes instruction relevance and subset diversity, but it provides no explicit mechanism for protecting indispensable evidence before diversity influences selection. Moreover, when relevance guidance is weak or spatially diffuse, the unified objective offers no separate stage for compensating through broader contextual selection.

We use \emph{contextual expansion} to denote the addition of visual tokens that are both informative and complementary to a protected set of query-critical evidence. This notion is stricter than diversity alone: a visually distinct but uninformative region may increase feature dispersion without providing useful contextual support. In AnchorPrune, contextual expansion is therefore performed only after a relevance anchor has been established.

\section{Method}
\label{sec:method}

\begin{figure*}[t]
\centering
\includegraphics[width=\textwidth]{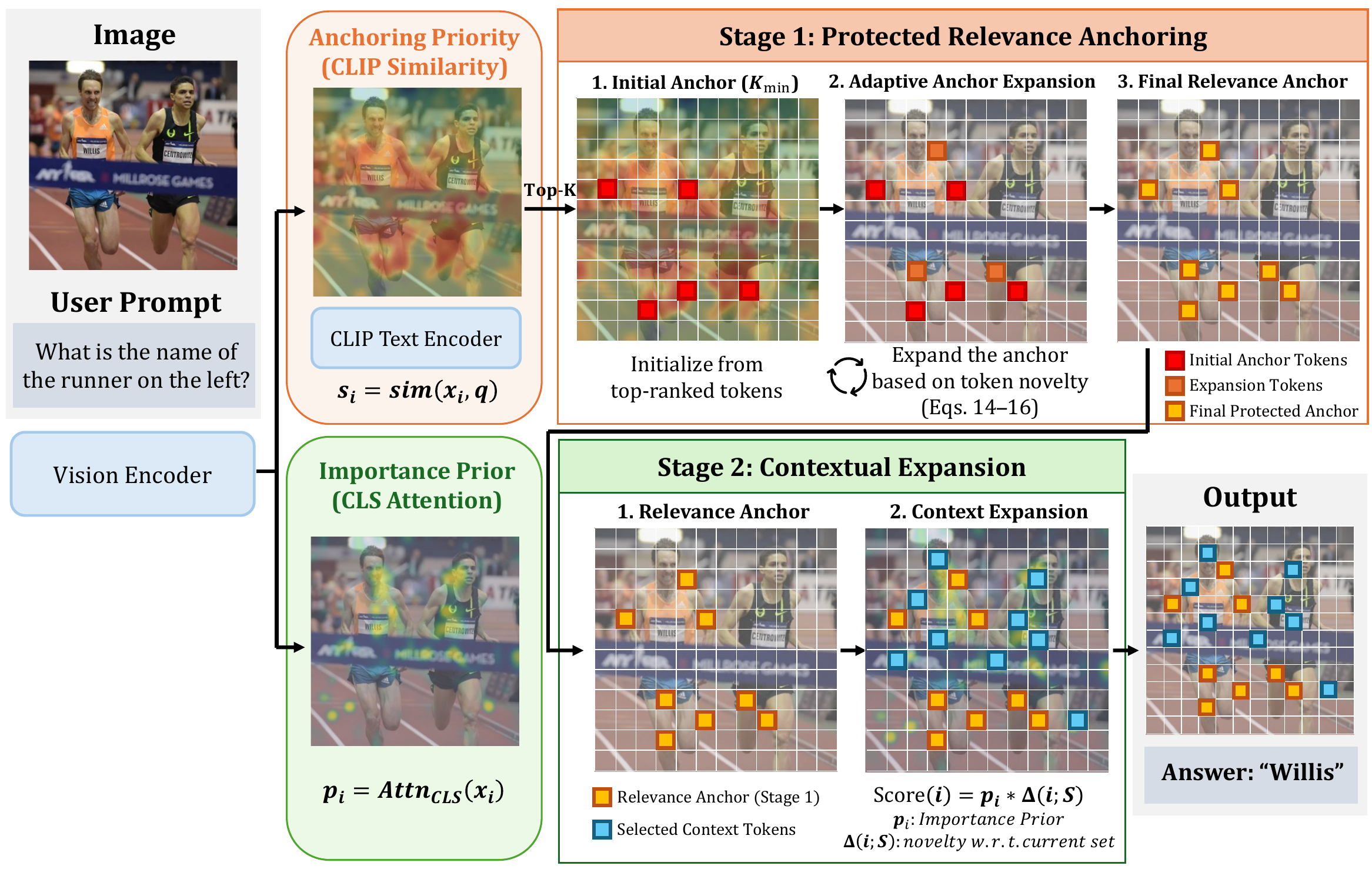}
\caption{\textbf{Overview of AnchorPrune.}
Given an image and an instruction, Stage~1 ranks visual tokens using an architecture-specific query-conditioned priority signal and constructs a protected relevance anchor whose size is determined adaptively from the novelty profile of the ranked tokens. Stage~2 initializes the retained set with this anchor and allocates the remaining budget through importance-weighted novelty, selecting tokens that are both globally informative and complementary to the evidence already retained. The selected tokens are restored to their native visual order and passed to the unchanged language model.}
\label{fig:method}
\end{figure*}

\subsection{Overview}
\label{sec:method_overview}

Figure~\ref{fig:method} provides an overview of AnchorPrune. AnchorPrune is a training-free visual token pruning framework that establishes a protected relevance anchor before expanding it with complementary visual context. Given visual tokens
$X=\{x_i\}_{i=1}^{N}$, a text query $q$, and a target token budget $K$, AnchorPrune constructs the retained set as
\begin{equation}
S
=
S_{\mathrm{rel}}\cup S_{\mathrm{ctx}},
\qquad
|S_{\mathrm{rel}}|=K_{\mathrm{rel}},
\qquad
|S_{\mathrm{ctx}}|=K_{\mathrm{ctx}},
\end{equation}
where $K_{\mathrm{rel}}+K_{\mathrm{ctx}}=K$.

The two subsets are selected sequentially. Stage~1 constructs a protected relevance anchor $S_{\mathrm{rel}}$ from tokens prioritized by the input instruction. Stage~2 uses the remaining budget to expand this anchor with tokens that are globally informative and non-redundant with respect to the current retained set. Because $S_{\mathrm{rel}}$ is fixed before contextual expansion, subsequent selections cannot displace query-critical evidence.

\subsection{Stage 1: Protected Relevance Anchoring}
\label{sec:relevance_anchoring}

Stage~1 assigns each visual token an anchoring-priority score $s_i$, where a larger value indicates higher priority for inclusion in the protected relevance anchor. The score is computed in an architecture-appropriate representation space in which visual and textual features are comparable.

\subsubsection*{CLIP-aligned VLMs.}
For models equipped with a paired CLIP vision--language encoder~\cite{clip}, we compute the priority score before the multimodal projector. Let
\begin{equation}
v_i
=
\frac{x_i}
{\lVert x_i\rVert_2}
\end{equation}
denote the normalized pre-projector visual embedding of token $i$, and let
$\{t_m\}_{m=1}^{M}$ denote normalized text embeddings produced by the corresponding text encoder. If the instruction exceeds the encoder context length, it is divided into $M$ segments. We define
\begin{equation}
s_i
=
-\frac{1}{M}
\sum_{m=1}^{M}
v_i^{\top}t_m .
\label{eq:negated_clip_similarity}
\end{equation}

Equation~\eqref{eq:negated_clip_similarity} defines the \emph{negated patch--text similarity}. Prior analysis found that raw CLIP local similarities can respond more strongly to background than to discriminative foreground regions, and that reversing these maps can improve localization~\cite{eclip}. We therefore use negated similarity as a training-free anchoring-priority signal, without interpreting it as calibrated semantic relevance. The two similarity directions are compared in Supplementary
Sec.~\ref{sec:clip_direction}.

\subsubsection*{Non-CLIP VLMs.}
For architectures without a comparable pre-projector vision--language space, we compute the priority score after the multimodal projector. Let
\begin{equation}
z_i=g_{\mathrm{mm}}(x_i)
\end{equation}
denote the projected visual token, and let $\{e_m\}_{m=1}^{M}$ denote the language-model embeddings of the instruction tokens. We define
\begin{equation}
s_i
=
\max_{1\leq m\leq M}
\left\langle
\frac{z_i}{\lVert z_i\rVert_2},
\frac{e_m}{\lVert e_m\rVert_2}
\right\rangle.
\label{eq:post_projector_relevance}
\end{equation}
Token-wise maximum matching preserves strong alignment with individual semantic components of the instruction, which can be obscured when all instruction embeddings are averaged into a single vector.

\subsubsection*{Relevance ranking.}
Using the architecture-specific score $s_i$, we sort the visual tokens in descending priority:
\begin{equation}
\pi=(\pi_1,\ldots,\pi_N),
\qquad
s_{\pi_1}\geq s_{\pi_2}\geq\cdots\geq s_{\pi_N}.
\end{equation}
For an anchor budget $K_{\mathrm{rel}}$, the protected relevance anchor is
\begin{equation}
S_{\mathrm{rel}}
=
\{\pi_1,\ldots,\pi_{K_{\mathrm{rel}}}\}.
\label{eq:relevance_anchor}
\end{equation}
No diversity penalty is applied in this stage. Consequently, correlated tokens may remain in the anchor when they jointly contain useful local evidence for answering the query.

\subsection{Adaptive Anchor Budget}
\label{sec:adaptive_budget}

A fixed relevance ratio is unnecessarily rigid because the structure of query-critical evidence varies across image--query pairs. In particular, the relevance-ranked sequence may contain different amounts of correlated query-related evidence. AnchorPrune therefore determines $K_{\mathrm{rel}}$ from the novelty profile of the relevance-ranked sequence.

We begin with a minimum anchor size $K_{\min}$ and impose an upper bound
\begin{equation}
K_{\max}
=
\left\lfloor \frac{K}{2}\right\rfloor ,
\end{equation}
ensuring that at least half of the total budget remains available for contextual expansion.

Let
\begin{equation}
A_{\min}
=
\{\pi_1,\ldots,\pi_{K_{\min}}\}
\end{equation}
denote the initial relevance anchor. Using visual features $h_i$, we measure the novelty of each subsequent relevance-ranked candidate $\pi_t$ relative to $A_{\min}$:
\begin{equation}
\Delta(\pi_t;A_{\min})
=
\min_{j\in A_{\min}}
\left(
1-
\frac{h_{\pi_t}^{\top}h_j}
{\lVert h_{\pi_t}\rVert_2\lVert h_j\rVert_2}
\right).
\label{eq:anchor_novelty}
\end{equation}
For $K_{\min}<t\leq K_{\max}$, we count the number of sufficiently novel candidates encountered after the initial anchor:
\begin{equation}
c_t
=
\sum_{u=K_{\min}+1}^{t}
\mathbf{1}
\left[
\Delta(\pi_u;A_{\min})>\tau
\right],
\end{equation}
where $\tau$ is a novelty threshold. Given a patience parameter $P$, the anchor budget is selected as
\begin{equation}
K_{\mathrm{rel}}
=
\begin{cases}
t, & \text{if $t$ is the first index satisfying } c_t \geq P, \\[2pt]
K_{\max}, & \text{otherwise}.
\end{cases}
\label{eq:adaptive_krel}
\end{equation}
The remaining context budget is
\begin{equation}
K_{\mathrm{ctx}}
=
K-K_{\mathrm{rel}}.
\end{equation}

When sufficiently novel candidates appear early in the relevance-ranked
sequence, suggesting that the initial query-related evidence has been
sufficiently covered, AnchorPrune terminates anchor construction at the
$P$-th novelty event. If fewer than $P$ such events are observed, it
conservatively retains the maximum anchor budget $K_{\max}$.

\subsection{Stage 2: Contextual Expansion}
\label{sec:context_expansion}

After constructing the protected relevance anchor, AnchorPrune allocates the remaining budget to contextual expansion. Unlike diversity-only sampling, Stage~2 does not favor novelty in isolation. Each candidate must be both globally informative and complementary to the evidence already retained.

\subsubsection*{Global importance prior.}
We assign each visual token a prior $p_i$ estimating its contribution to the global visual representation. The prior is instantiated according to the vision-encoder architecture.

For encoders with a dedicated \texttt{[CLS]} token, let
$A_{h,\mathrm{cls},i}$ denote the attention from the \texttt{[CLS]} token to visual token $i$ at attention head $h$. We compute
\begin{equation}
p_i
=
\frac{1}{H}
\sum_{h=1}^{H}
A_{h,\mathrm{cls},i}.
\label{eq:cls_prior}
\end{equation}

For encoders without a dedicated global token, we use the average attention mass received by token $i$:
\begin{equation}
\widetilde{p}_i
=
\frac{1}{HN}
\sum_{h=1}^{H}
\sum_{j=1}^{N}
A_{h,j,i}.
\end{equation}
If the architecture merges multiple encoder tokens into a single candidate token, let $G(i)$ denote the corresponding pre-merge token group. The aligned prior is
\begin{equation}
p_i
=
\frac{1}{|G(i)|}
\sum_{j\in G(i)}
\widetilde{p}_j .
\label{eq:merged_prior}
\end{equation}
When no merging is applied, $G(i)=\{i\}$.

\subsubsection*{Anchor-conditioned novelty.}
We initialize the retained set with the protected relevance anchor:
\begin{equation}
S\leftarrow S_{\mathrm{rel}}.
\end{equation}
For candidate token $i\notin S$, its novelty relative to the current retained set is
\begin{equation}
\Delta(i;S)
=
\min_{j\in S}
\left(
1-
\frac{h_i^{\top}h_j}
{\lVert h_i\rVert_2\lVert h_j\rVert_2}
\right).
\label{eq:context_novelty}
\end{equation}
Because $S$ is initialized with $S_{\mathrm{rel}}$, novelty is measured relative to both the protected anchor and the contextual tokens selected in preceding iterations.

\subsubsection*{Importance-weighted expansion.}
At each iteration, AnchorPrune selects
\begin{equation}
i^{\star}
=
\arg\max_{i\notin S}
p_i\,\Delta(i;S),
\qquad
S\leftarrow S\cup\{i^{\star}\}
\label{eq:importance_weighted_novelty}
\end{equation}

until $|S|=K$. The resulting context subset is
\begin{equation}
S_{\mathrm{ctx}}
=
S\setminus S_{\mathrm{rel}}.
\end{equation}

The multiplicative objective in Eq.~\eqref{eq:importance_weighted_novelty} suppresses both visually distinct but uninformative tokens and globally important but redundant ones. Each selected token must therefore be simultaneously informative and complementary to the current retained set, producing contextual expansion rather than feature dispersion alone. The selected tokens are restored to their native visual order and passed to the unchanged language model without retraining or architectural modification.

\section{Experiments}
\label{sec:experiments}

We evaluate AnchorPrune in terms of performance preservation under aggressive visual-token compression, generalization across heterogeneous image and video VLM architectures, and the effectiveness of relevance-anchored contextual expansion. Our evaluation spans two CLIP-aligned image VLMs with substantially different visual-sequence lengths, a non-CLIP image VLM, and a video VLM. Detailed implementation settings, video breakdowns, and additional ablations
are provided in Supplementary Secs.~\ref{sec:supp_implementation},
\ref{sec:supp_video_results}, and~\ref{sec:supp_ablations}, respectively.

\subsection{Experimental Setup}
\label{sec:experimental_setup}

\subsubsection*{Models and evaluation.}
We evaluate AnchorPrune on LLaVA-1.5-7B~\cite{llava}, 
LLaVA-NeXT-7B~\cite{llavanext},
Qwen2.5-VL-7B~\cite{qwen}, and
LLaVA-Video-7B~\cite{llava_video}.

For LLaVA-1.5-7B, we retain 128, 64, or 32 tokens from 576 visual tokens.
For LLaVA-NeXT-7B, we retain 640, 320, or 160 tokens from 2,880 visual tokens.
Both models are evaluated on VQAv2~\cite{vqav2},
TextVQA~\cite{textvqa}, GQA~\cite{gqa},
ScienceQA-IMG~\cite{scienceqa}, MME~\cite{mme},
POPE~\cite{pope}, MMBench-EN/CN~\cite{mmbench},
and MM-Vet~\cite{mmvet}.

For Qwen2.5-VL-7B, we retain 256, 128, or 64 tokens from
1,296 visual tokens and evaluate on MME, TextVQA,
DocVQA~\cite{docvqa}, AI2D~\cite{ai2d},
MMMU~\cite{mmmu}, and MMBench-EN/CN.

For LLaVA-Video-7B, we evaluate 16-frame inputs and retain
1,024 or 512 tokens from 2,704 visual tokens on
Video-MME~\cite{video_mme}, EgoSchema~\cite{egoschema},
and TempCompass~\cite{tempcompass}.

Results for all evaluated model--budget settings are reported in Tables~\ref{tab:llava15_main}, \ref{tab:llava_next_main}, \ref{tab:qwen_main}, and~\ref{tab:video_compact}. We additionally conduct a controlled Stage-2 ablation on LLaVA-1.5-7B.

\subsubsection*{Baselines and protocol.}
We compare against representative training-free methods based on attention or saliency, progressive token reduction, token merging, diversity, and unified relevance--diversity selection~\cite{fastv,pyramiddrop,sparsevlm,prumerge,visionzip,divprune,cdpruner}.
All methods use matched model checkpoints, input processing, decoding settings, benchmark protocols, and retained-token budgets.
We use \texttt{lmms-eval}~\cite{lmms_eval} whenever supported.
Architecture-specific implementation details, hyperparameters,
and hardware configurations are provided in Supplementary Sec.~\ref{sec:supp_implementation}.

\subsubsection*{Relative retained performance.}
To aggregate benchmarks with different metric scales, we report the average performance retained relative to the unpruned backbone:
\begin{equation}
    \mathrm{Rel.}(m)
    =
    \frac{1}{|\mathcal{B}|}
    \sum_{b\in\mathcal{B}}
    \frac{\mathrm{Score}_{b}(m)}
         {\mathrm{Score}_{b}(\mathrm{Full})}
    \times 100,
    \label{eq:relative_performance}
\end{equation}
where $\mathcal{B}$ denotes the set of metrics reported in the corresponding table.

\begin{table*}[t]
\centering
\caption{\textbf{Main results on LLaVA-1.5-7B across visual-token budgets.}
We evaluate AnchorPrune on a compact CLIP-aligned VLM containing 576 visual tokens.
Best and second-best \emph{Rel.} values within each budget are highlighted in \textbf{bold} and \underline{underlined}, respectively.}
\label{tab:llava15_main}
\setlength{\tabcolsep}{3.6pt}
\renewcommand{\arraystretch}{1.05}
\resizebox{\textwidth}{!}{
\begin{tabular}{lccccccccc|c}
\toprule
Method & VQAv2 & TextVQA & GQA & SQA$^{\mathrm{IMG}}$ & MME & POPE &
MMB$^{\mathrm{EN}}$ & MMB$^{\mathrm{CN}}$ & MM-Vet & Rel. \\
\midrule
\multicolumn{11}{c}{\textbf{Upper Bound: All 576 Tokens (100\%)}} \\
\midrule
LLaVA-1.5-7B & 76.7 & 58.3 & 61.9 & 69.6 & 1509.1 & 85.8 & 64.0 & 55.6 & 31.3 & 100.0 \\
\midrule
\multicolumn{11}{c}{\textbf{Retain 128 Tokens} ($\downarrow$ 77.8\%)} \\
\midrule
FastV (ECCV'24) & 72.7 & 56.4 & 56.9 & 68.9 & 1456.1 & 76.2 & 62.5 & 53.6 & 28.2 & 94.7 \\
PyramidDrop (CVPR'25) & 73.4 & 55.9 & 56.5 & 69.2 & 1382.5 & 78.8 & 62.7 & 51.8 & 26.3 & 93.4 \\
SparseVLM (ICML'25) & 74.4 & 56.7 & 58.6 & 68.7 & 1435.6 & 85.0 & 63.5 & 54.4 & 30.4 & \second{97.3} \\
PruMerge+ (ICCV'25) & 72.3 & 54.7 & 57.7 & 69.5 & 1406.3 & 81.0 & 60.1 & 51.4 & 26.6 & 93.3 \\
VisionZip (CVPR'25) & 73.8 & 56.8 & 57.5 & 68.7 & 1435.1 & 83.1 & 62.4 & 53.5 & 32.7 & \second{97.3} \\
DivPrune (CVPR'25) & 74.2 & 56.4 & 59.2 & 69.0 & 1405.3 & 86.8 & 62.6 & 52.5 & 29.2 & 96.5 \\
CDPruner (NeurIPS'25) & 75.0 & 56.3 & 59.6 & 68.6 & 1413.3 & 87.1 & 62.4 & 51.9 & 28.1 & 96.1 \\
\ourrow \ours & 75.1 & 57.0 & 59.8 & 68.9 & 1454.3 & 86.8 & 63.3 & 53.2 & 32.8 & \best{98.7} \\
\midrule
\multicolumn{11}{c}{\textbf{Retain 64 Tokens} ($\downarrow$ 88.9\%)} \\
\midrule
FastV (ECCV'24) & 68.2 & 54.1 & 53.6 & 69.3 & 1357.8 & 68.1 & 60.1 & 50.3 & 26.4 & 89.5 \\
PyramidDrop (CVPR'25) & 69.0 & 53.0 & 52.8 & 68.1 & 1204.7 & 69.3 & 58.2 & 45.5 & 20.2 & 84.7 \\
SparseVLM (ICML'25) & 68.7 & 53.4 & 53.7 & 69.4 & 1318.7 & 77.5 & 59.9 & 48.9 & 23.6 & 89.1 \\
PruMerge+ (ICCV'25) & 69.7 & 54.1 & 55.3 & 69.5 & 1318.7 & 74.9 & 58.8 & 49.6 & 24.1 & 89.5 \\
VisionZip (CVPR'25) & 70.6 & 55.4 & 55.0 & 69.0 & 1373.1 & 77.1 & 60.0 & 51.1 & 30.0 & 93.0 \\
DivPrune (CVPR'25) & 72.5 & 54.8 & 57.8 & 68.3 & 1343.8 & 85.5 & 59.2 & 49.4 & 24.2 & 91.9 \\
CDPruner (NeurIPS'25) & 73.9 & 55.5 & 58.9 & 68.4 & 1400.6 & 87.3 & 61.2 & 49.8 & 26.1 & \second{94.2} \\
\ourrow \ours & 73.8 & 56.1 & 58.6 & 68.8 & 1420.2 & 85.8 & 61.3 & 52.0 & 30.1 & \best{96.2} \\
\midrule
\multicolumn{11}{c}{\textbf{Retain 32 Tokens} ($\downarrow$ 94.4\%)} \\
\midrule
PruMerge+ (ICCV'25) & 65.5 & 52.0 & 53.3 & 69.2 & 1227.0 & 69.6 & 56.4 & 45.2 & 22.1 & 84.7 \\
VisionZip (CVPR'25) & 65.7 & 53.2 & 51.8 & 68.8 & 1255.8 & 68.8 & 57.2 & 48.0 & 24.0 & 86.1 \\
DivPrune (CVPR'25) & 69.4 & 53.0 & 54.7 & 68.6 & 1267.9 & 81.9 & 57.9 & 45.8 & 24.3 & 88.7 \\
CDPruner (NeurIPS'25) & 72.1 & 52.9 & 57.2 & 69.1 & 1362.6 & 87.5 & 60.1 & 46.0 & 27.7 & \second{92.6} \\
\ourrow \ours & 71.7 & 54.2 & 56.9 & 69.5 & 1394.7 & 85.3 & 60.1 & 49.9 & 28.7 & \best{93.9} \\
\bottomrule
\end{tabular}}
\end{table*}

\subsection{Main Results}
\label{sec:main_results}

\subsubsection*{CLIP-aligned image VLM.}
Table~\ref{tab:llava15_main} reports results on LLaVA-1.5-7B, whose visual representation contains 576 tokens.
This compact setting provides a demanding test because the original sequence contains substantially less redundancy than high-resolution multi-crop representations.

AnchorPrune achieves the highest aggregate retention at all three budgets. It retains 98.7\% and 96.2\% of the full-token baseline with 128 and 64 tokens, respectively. Under the most aggressive setting, it retains 93.9\% using only 32 visual tokens, exceeding the second-best method by 1.3 percentage points.

The gains are distributed across benchmarks rather than being driven by a single metric. At 32 tokens, AnchorPrune is particularly strong on MME, TextVQA, MMBench-CN, and MM-Vet, showing that relevance-anchored contextual expansion remains effective even when the original visual sequence is comparatively compact.

\begin{table*}[t]
\centering
\caption{\textbf{Main results on LLaVA-NeXT-7B across visual-token budgets.}
We evaluate scalability to a high-resolution multi-crop representation containing 2,880 visual tokens.
Best and second-best \emph{Rel.} values within each budget are highlighted in \textbf{bold} and \underline{underlined}, respectively.}
\label{tab:llava_next_main}
\setlength{\tabcolsep}{3.0pt}
\renewcommand{\arraystretch}{1.05}
\resizebox{\textwidth}{!}{
\begin{tabular}{lccccccccc|c}
\toprule
Method & VQAv2 & TextVQA & GQA & SQA$^{\mathrm{IMG}}$ & MME & POPE &
MMB$^{\mathrm{EN}}$ & MMB$^{\mathrm{CN}}$ & MM-Vet & Rel. \\
\midrule
\multicolumn{11}{c}{\textbf{Upper Bound: All 2,880 Tokens (100\%)}} \\
\midrule
LLaVA-NeXT-7B & 80.2 & 61.3 & 64.2 & 70.2 & 1528.8 & 86.4 & 67.2 & 56.5 & 39.3 & 100.0 \\
\midrule
\multicolumn{11}{c}{\textbf{Retain 640 Tokens} ($\downarrow$ 77.8\%)} \\
\midrule
FastV (ECCV'24) & 77.2 & 57.6 & 60.5 & 67.9 & 1513.6 & 82.6 & 62.5 & 54.0 & 25.5 & 92.1 \\
PyramidDrop (CVPR'25) & 78.7 & 59.8 & 61.0 & 68.4 & 1470.3 & 84.6 & 65.3 & 57.3 & 26.6 & 94.3 \\
SparseVLM (ICML'25) & 77.1 & 57.8 & 59.4 & 67.8 & 1481.7 & 84.7 & 64.9 & 56.9 & 32.6 & 95.0 \\
PruMerge+ (ICCV'25) & 70.5 & 51.0 & 55.4 & 70.0 & 1295.0 & 67.1 & 62.8 & 55.2 & 31.9 & 88.0 \\
VisionZip (CVPR'25) & 77.3 & 60.1 & 61.3 & 68.2 & 1461.5 & 86.0 & 64.6 & 56.3 & 35.8 & 96.6 \\
DivPrune (CVPR'25) & 78.9 & 52.6 & 58.1 & 67.4 & 1314.6 & 79.6 & 62.8 & 53.8 & 28.8 & 90.1 \\
CDPruner (NeurIPS'25) & 78.2 & 58.8 & 62.5 & 68.2 & 1491.3 & 87.4 & 65.9 & 56.6 & 33.5 & \second{96.7} \\
\ourrow \ours & 80.2 & 61.3 & 64.3 & 70.3 & 1528.8 & 86.4 & 67.1 & 56.5 & 37.0 & \best{99.4} \\
\midrule
\multicolumn{11}{c}{\textbf{Retain 320 Tokens} ($\downarrow$ 88.9\%)} \\
\midrule
FastV (ECCV'24) & 71.0 & 54.1 & 55.7 & 67.0 & 1327.2 & 73.9 & 60.2 & 50.8 & 21.7 & 85.1 \\
PyramidDrop (CVPR'25) & 76.2 & 56.8 & 57.9 & 67.7 & 1451.6 & 77.8 & 64.2 & 54.3 & 29.3 & 91.7 \\
SparseVLM (ICML'25) & 73.4 & 55.5 & 57.4 & 67.6 & 1419.1 & 81.2 & 63.2 & 54.0 & 29.8 & 91.1 \\
PruMerge+ (ICCV'25) & 67.7 & 49.6 & 53.7 & 69.6 & 1221.2 & 61.2 & 61.6 & 52.9 & 25.4 & 83.2 \\
VisionZip (CVPR'25) & 74.4 & 58.9 & 58.9 & 67.6 & 1410.1 & 82.2 & 63.1 & 55.0 & 31.5 & 92.9 \\
DivPrune (CVPR'25) & 76.6 & 50.8 & 56.1 & 67.3 & 1277.6 & 74.4 & 60.4 & 50.4 & 26.7 & 86.5 \\
CDPruner (NeurIPS'25) & 76.8 & 57.3 & 61.3 & 67.2 & 1482.9 & 87.6 & 64.6 & 55.2 & 35.7 & \second{95.9} \\
\ourrow \ours & 79.4 & 59.6 & 62.9 & 69.0 & 1507.5 & 87.2 & 65.3 & 56.9 & 37.2 & \best{98.3} \\
\midrule
\multicolumn{11}{c}{\textbf{Retain 160 Tokens} ($\downarrow$ 94.4\%)} \\
\midrule
PruMerge+ (ICCV'25) & 63.7 & 47.7 & 51.8 & 70.2 & 1153.1 & 56.8 & 58.0 & 47.3 & 27.3 & 79.8 \\
VisionZip (CVPR'25) & 70.0 & 56.0 & 55.3 & 67.8 & 1326.3 & 74.8 & 58.8 & 51.8 & 26.7 & 86.9 \\
DivPrune (CVPR'25) & 73.9 & 48.1 & 53.2 & 67.3 & 1211.7 & 67.9 & 57.4 & 47.1 & 19.7 & 80.7 \\
CDPruner (NeurIPS'25) & 75.2 & 55.7 & 60.8 & 67.4 & 1430.8 & 86.7 & 64.2 & 54.0 & 29.9 & \second{92.9} \\
\ourrow \ours & 78.6 & 60.1 & 62.6 & 68.7 & 1481.1 & 87.5 & 65.5 & 56.9 & 35.8 & \best{97.6} \\
\bottomrule
\end{tabular}}
\end{table*}

\subsubsection*{Scaling to high-resolution visual sequences.}
Table~\ref{tab:llava_next_main} evaluates AnchorPrune on LLaVA-NeXT-7B, whose high-resolution multi-crop representation contains 2,880 visual tokens.
This setting tests whether the same selection principle scales to substantially longer and more redundant visual sequences.

AnchorPrune achieves the highest retained performance at every evaluated budget. With 640 tokens, it retains 99.4\% of the full-token baseline, outperforming the second-best method by 2.7 percentage points in \emph{Rel.} At 320 tokens, AnchorPrune achieves 98.3\%, compared with 95.9\% for the strongest competing method.

The advantage becomes larger under the most aggressive setting. With only 160 of 2,880 tokens retained, AnchorPrune preserves 97.6\% of the full-token baseline, exceeding the second-best result by 4.7 percentage points. It also remains close to the unpruned model on TextVQA, MME, POPE, and both MMBench variants. Together with the LLaVA-1.5 results, this demonstrates that relevance-anchored contextual expansion is effective across both compact and high-resolution CLIP-aligned visual sequences.

\begin{table*}[t]
\centering
\caption{\textbf{Main results on Qwen2.5-VL-7B across visual-token budgets.}
We evaluate cross-architecture generalization on a non-CLIP VLM under matched inference settings.
Best and second-best \emph{Rel.} values within each budget are highlighted in \textbf{bold} and \underline{underlined}, respectively.}
\label{tab:qwen_main}
\setlength{\tabcolsep}{3.0pt}
\renewcommand{\arraystretch}{1.05}
\resizebox{\textwidth}{!}{
\begin{tabular}{lccccccc|c}
\toprule
Method & MME & TextVQA & DocVQA & AI2D & MMMU &
MMB$^{\mathrm{EN}}$ & MMB$^{\mathrm{CN}}$ & Rel. \\
\midrule
\multicolumn{9}{c}{\textbf{Upper Bound: All 1,296 Tokens (100\%)}} \\
\midrule
Qwen2.5-VL-7B & 1687.9 & 77.3 & 94.3 & 83.2 & 50.6 & 83.1 & 80.5 & 100.0 \\
\midrule
\multicolumn{9}{c}{\textbf{Retain 256 Tokens} ($\downarrow$ 80.2\%)} \\
\midrule
FastV (ECCV'24) & 1638.9 & 74.9 & 60.0 & 78.3 & 49.3 & 79.8 & 77.3 & 91.6 \\
DivPrune (CVPR'25) & 1671.9 & 70.0 & 67.9 & 77.8 & 48.6 & 80.0 & 77.2 & \second{91.9} \\
CDPruner (NeurIPS'25) & 1621.9 & 63.2 & 52.7 & 78.0 & 47.7 & 78.0 & 77.0 & 87.3 \\
\ourrow \ours & 1657.5 & 72.6 & 72.9 & 78.8 & 48.6 & 80.8 & 78.2 & \best{93.5} \\
\midrule
\multicolumn{9}{c}{\textbf{Retain 128 Tokens} ($\downarrow$ 90.1\%)} \\
\midrule
FastV (ECCV'24) & 1554.8 & 70.7 & 40.0 & 72.9 & 47.1 & 76.3 & 72.1 & 84.0 \\
DivPrune (CVPR'25) & 1642.4 & 66.3 & 49.8 & 75.2 & 47.3 & 76.9 & 75.4 & \second{86.6} \\
CDPruner (NeurIPS'25) & 1572.1 & 57.7 & 36.3 & 76.9 & 45.9 & 77.9 & 74.7 & 82.3 \\
\ourrow \ours & 1663.9 & 68.4 & 51.0 & 77.7 & 46.8 & 79.6 & 75.7 & \best{88.1} \\
\midrule
\multicolumn{9}{c}{\textbf{Retain 64 Tokens} ($\downarrow$ 95.1\%)} \\
\midrule
FastV (ECCV'24) & 1381.4 & 63.7 & 27.7 & 68.4 & 45.4 & 67.5 & 64.7 & 75.3 \\
DivPrune (CVPR'25) & 1544.1 & 60.7 & 31.6 & 70.7 & 46.9 & 74.6 & 71.7 & \second{80.0} \\
CDPruner (NeurIPS'25) & 1431.7 & 51.6 & 22.8 & 73.6 & 44.7 & 73.9 & 72.9 & 76.0 \\
\ourrow \ours & 1563.2 & 61.5 & 31.3 & 74.2 & 45.2 & 75.7 & 73.2 & \best{80.8} \\
\bottomrule
\end{tabular}}
\end{table*}

\subsubsection*{Generalization to a non-CLIP architecture.}
Table~\ref{tab:qwen_main} evaluates AnchorPrune on Qwen2.5-VL-7B.
Unlike the LLaVA models, Qwen2.5-VL does not expose the same paired pre-projector CLIP representation.
AnchorPrune therefore computes its Stage-1 priority in the post-projector embedding space while preserving the same protected-anchor design.

AnchorPrune achieves the highest \emph{Rel.} at all three budgets. At 256 tokens, it retains 93.5\% of the full-token baseline, improving over the second-best result by 1.6 percentage points. At 128 tokens, AnchorPrune obtains 88.1\%, compared with 86.6\% for the strongest baseline. Even when only 64 of 1,296 tokens remain, AnchorPrune achieves the best overall result at 80.8\%.

The improvements are consistent across general, text-oriented, and document-oriented benchmarks, indicating that the anchoring principle transfers beyond CLIP-aligned architectures. These results demonstrate that AnchorPrune is not tied to a specific visual encoder or relevance representation.

\begin{table}[t]
\centering
\caption{\textbf{Video results on LLaVA-Video-7B.}
We report aggregate performance under 16-frame evaluation.
Detailed short-, medium-, and long-video results are provided in
Supplementary Sec.~\ref{sec:supp_video_results}.
Best and second-best \emph{Rel.} values within each budget are highlighted in \textbf{bold} and \underline{underlined}, respectively.}
\label{tab:video_compact}
\setlength{\tabcolsep}{2.4pt}
\renewcommand{\arraystretch}{1.02}
\resizebox{\columnwidth}{!}{
\begin{tabular}{lcccc|c}
\toprule
Method &
\shortstack{Video-MME\\w/o Subtitles} &
\shortstack{Video-MME\\w/ Subtitles} &
\shortstack{EgoSchema\\500-subset} &
\shortstack{TempCompass\\Avg.} &
Rel. \\
\midrule
\multicolumn{6}{c}{\textbf{Upper Bound: All 2,704 Visual Tokens}} \\
\midrule
LLaVA-Video-7B & 59.96 & 61.33 & 55.4 & 67.1 & 100.0 \\
\midrule
\multicolumn{6}{c}{\textbf{Retain 1,024 Tokens} ($\downarrow$ 62.1\%)} \\
\midrule
FastV (ECCV'24) & 59.26 & 60.59 & 52.8 & 65.6 & \second{97.7} \\
DivPrune (CVPR'25) & 57.00 & 59.15 & 53.2 & 64.9 & 96.1 \\
CDPruner (NeurIPS'25) & 57.85 & 59.81 & 52.6 & 65.6 & 96.8 \\
\ourrow \ours & 58.19 & 60.89 & 55.0 & 65.4 & \best{98.3} \\
\midrule
\multicolumn{6}{c}{\textbf{Retain 512 Tokens} ($\downarrow$ 81.1\%)} \\
\midrule
FastV (ECCV'24) & 56.37 & 57.63 & 50.2 & 61.1 & 92.4 \\
DivPrune (CVPR'25) & 55.89 & 58.19 & 52.0 & 61.3 & \second{93.4} \\
CDPruner (NeurIPS'25) & 54.22 & 57.00 & 51.4 & 60.6 & 91.7 \\
\ourrow \ours & 56.26 & 58.07 & 53.0 & 61.8 & \best{94.1} \\
\bottomrule
\end{tabular}}
\end{table}

\subsubsection*{Video Generalization.}
% \paragraph{Video generalization.}
Table~\ref{tab:video_compact} evaluates AnchorPrune on LLaVA-Video-7B under 16-frame inputs.
AnchorPrune achieves the highest aggregate retained performance at both budgets, preserving 98.3\% and 94.1\% of the full-token baseline with 1,024 and 512 tokens, respectively. This result indicates that relevance-anchored contextual expansion extends to temporally structured visual inputs.

\subsubsection*{Overall trend.}
Across compact and high-resolution CLIP-aligned image VLMs, a non-CLIP image VLM, and a video VLM, AnchorPrune consistently achieves the strongest retained performance.
Its advantage generally becomes clearer as the token budget decreases, supporting the principle that query-critical evidence should be protected before complementary context is expanded.

\subsection{Efficiency Analysis}
\label{sec:efficiency}

\begin{figure*}[t]
    \centering
    \includegraphics[width=0.96\textwidth]{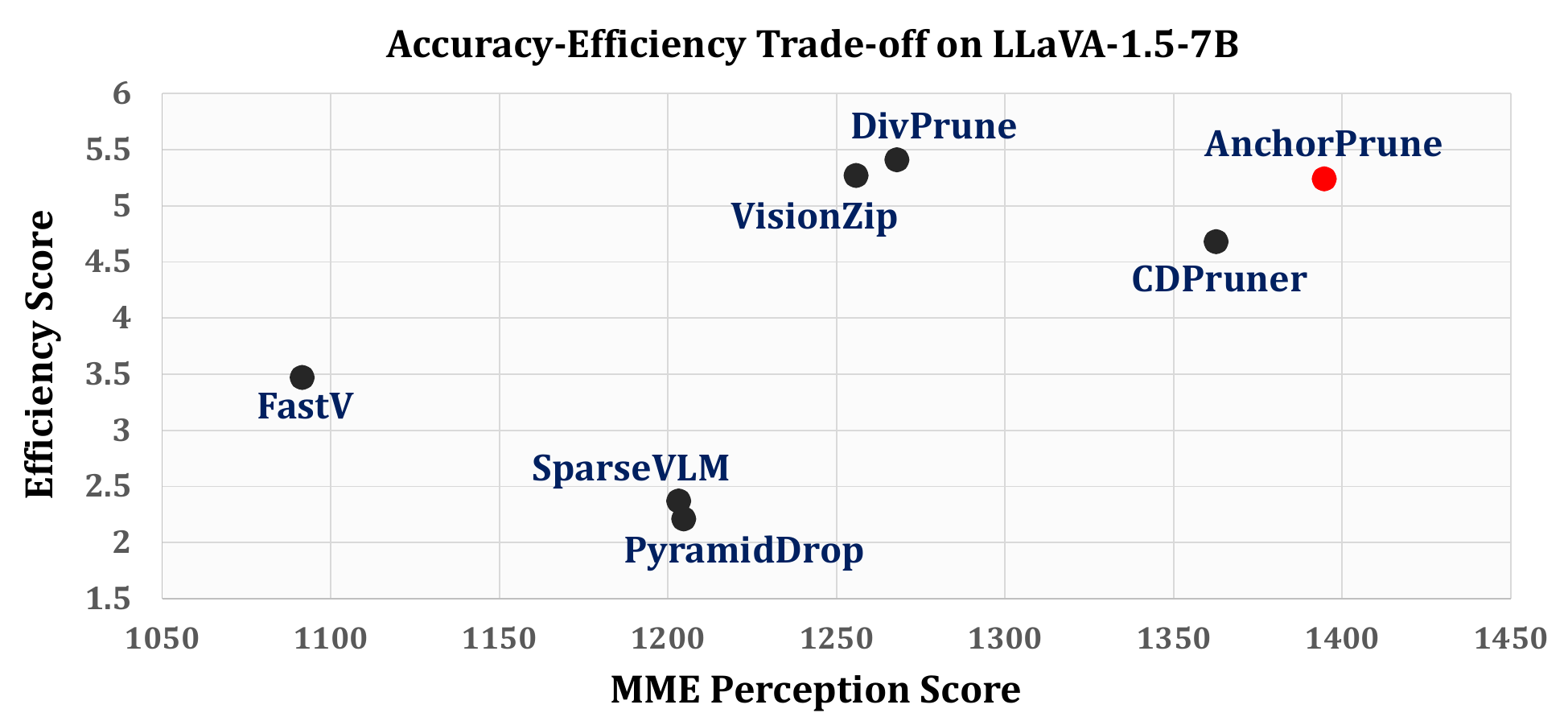}
    \caption{\textbf{Accuracy--efficiency trade-off on LLaVA-1.5-7B.}
    The efficiency score jointly reflects prefill speedup and resident GPU memory relative to the full-token baseline.
    AnchorPrune occupies the favorable high-accuracy, high-efficiency region, combining the strongest task performance with competitive practical efficiency.}
    \label{fig:efficiency_tradeoff}
\end{figure*}

Finally, we evaluate the practical efficiency of AnchorPrune, including token-selection overhead. To summarize the joint trade-off between prefill latency and resident memory in a single quantity, we define an \emph{Efficiency Score} as
\begin{equation}
    \mathrm{Efficiency\ Score}
    =
    \mathrm{Prefill\ Speedup}
    \times
    \frac{\mathrm{Memory}_{\mathrm{Full}}}
         {\mathrm{Memory}_{\mathrm{Method}}},
    \label{eq:efficiency_score}
\end{equation}
where \(\mathrm{Prefill\ Speedup}\) denotes the speedup relative to the unpruned model,
\(\mathrm{Memory}_{\mathrm{Full}}\) is the resident GPU memory usage of the full-token baseline,
and \(\mathrm{Memory}_{\mathrm{Method}}\) is the resident GPU memory usage of the pruned method.

Figure~\ref{fig:efficiency_tradeoff} plots this score against MME performance under a matched budget of 32 retained visual tokens. AnchorPrune achieves the highest MME score among the evaluated pruning methods while maintaining a competitive efficiency score of \(5.24\). VisionZip and DivPrune obtain slightly higher efficiency scores of \(5.28\) and \(5.41\), respectively, due to marginally lower resident memory, but their MME scores are substantially lower. CDPruner reaches a similar prefill speedup but has a lower efficiency score of \(4.68\) because of its larger resident memory footprint. These results show that AnchorPrune provides the strongest accuracy--efficiency operating point rather than improving accuracy through a costly selection mechanism.

This result is consistent with its lightweight design: Stage~1 uses simple anchoring-priority scoring, while Stage~2 performs greedy importance-weighted expansion without requiring kernel-based subset optimization. Detailed FLOPs, latency, memory, and performance measurements are provided
in Supplementary Sec.~\ref{sec:supp_efficiency}.

\subsection{Ablation Study}
\label{sec:ablation}

\begin{table}[t]
\centering
\caption{\textbf{Controlled Stage-2 ablation on LLaVA-1.5-7B.}
All variants use the same protected relevance anchor and differ only in how the remaining budget is allocated.
Best and second-best \emph{Rel.} values within each budget are highlighted in \textbf{bold} and \underline{underlined}, respectively.}
\label{tab:stage2_ablation}
\setlength{\tabcolsep}{3.0pt}
\renewcommand{\arraystretch}{1.0}
\resizebox{\columnwidth}{!}{
\begin{tabular}{lccccc|c}
\toprule
Method & MME & ChartQA & DocVQA & TextVQA & MMB$^{\mathrm{CN}}$ & Rel. \\
\midrule
\multicolumn{7}{c}{\textbf{Upper Bound: All 576 Tokens (100\%)}} \\
\midrule
LLaVA-1.5-7B & 1509.1 & 18.2 & 21.5 & 58.3 & 55.6 & 100.0 \\
\midrule
\multicolumn{7}{c}{\textbf{Retain 64 Tokens}} \\
\midrule
Diversity Only & 1391.0 & 15.1 & 15.5 & 54.5 & 48.0 & 85.4 \\
Additive Relevance--Diversity & 1380.9 & 15.9 & 16.4 & 54.8 & 48.5 & \second{87.3} \\
Multiplicative Relevance--Diversity & 1379.9 & 15.9 & 16.6 & 54.5 & 48.2 & 87.2 \\
\ourrow Importance-Weighted Contextual Expansion \textbf{(Ours)} &
1420.2 & 16.7 & 17.1 & 56.1 & 52.0 & \best{91.0} \\
\midrule
\multicolumn{7}{c}{\textbf{Retain 32 Tokens}} \\
\midrule
Diversity Only & 1317.9 & 14.6 & 12.8 & 52.6 & 44.2 & 79.4 \\
Additive Relevance--Diversity & 1344.5 & 14.2 & 14.2 & 52.7 & 45.2 & 81.0 \\
Multiplicative Relevance--Diversity & 1329.4 & 14.4 & 14.4 & 52.7 & 45.0 & \second{81.1} \\
\ourrow Importance-Weighted Contextual Expansion \textbf{(Ours)} &
1394.7 & 15.1 & 14.5 & 54.2 & 49.9 & \best{85.1} \\
\bottomrule
\end{tabular}}
\end{table}

\subsubsection*{Stage-2 expansion strategy.}
We isolate the contribution of the Stage-2 selection criterion in Table~\ref{tab:stage2_ablation}.
All variants share the same protected Stage-1 relevance anchor and differ only in the criterion used to allocate the remaining token budget.
We compare diversity-only selection, additive relevance--diversity coupling, multiplicative relevance--diversity coupling, and the proposed importance-weighted contextual expansion rule.
For this ablation, \emph{Rel.} is computed over
MME~\cite{mme}, ChartQA~\cite{chartqa},
DocVQA~\cite{docvqa}, TextVQA~\cite{textvqa},
and MMBench-CN~\cite{mmbench}, as reported in the table.

At 64 retained tokens, diversity-only selection obtains 85.4\% \emph{Rel.} The additive and multiplicative relevance--diversity formulations improve \emph{Rel.} to 87.3\% and 87.2\%, respectively. In contrast, importance-weighted contextual expansion achieves 91.0\%, outperforming the strongest alternative by 3.7 percentage points.

The same conclusion holds under stronger compression.
At 32 retained tokens, the proposed rule achieves 85.1\% \emph{Rel.}, compared with 81.1\% for the strongest alternative and 79.4\% for diversity-only selection.
The gains are especially pronounced on MME, TextVQA, and MMBench-CN.

These results show that contextual expansion is not equivalent to diversity maximization.
Diversity-only selection may allocate capacity to visually distinct but weakly informative regions, while direct relevance--diversity coupling permits strength in one criterion to compensate for weakness in the other.
Starting from the same protected relevance anchor, importance-weighted contextual expansion selects tokens that are simultaneously informative and complementary to the evidence already retained.
\section{Conclusion}
\label{sec:conclusion}

We presented \textbf{AnchorPrune}, a training-free visual token pruning framework that constructs an adaptive protected relevance anchor before expanding it through importance-weighted novelty. This sequential design preserves query-critical evidence while adding informative, non-redundant context. Across CLIP-aligned and non-CLIP image and video VLMs, AnchorPrune consistently improves performance preservation under matched token budgets, with the largest gains under severe compression. On LLaVA-NeXT-7B, it preserves 97.6\% of full-token performance using only 160 of 2,880 visual tokens. These results establish relevance-anchored contextual expansion as an effective principle for efficient multimodal inference.

\subsubsection*{Acknowledgements.}
 This research was supported by the Electronics and Telecommunications Research Institute (ETRI) Grant funded by the Korean Government (Fundamental Technology Research for Human-Centric Autonomous Intelligent Systems) under Grant 25ZB1200.
\bibliographystyle{splncs04}
\bibliography{main}

@inproceedings{llava,
  title={Improved baselines with visual instruction tuning},
  author={Liu, Haotian and Li, Chunyuan and Li, Yuheng and Lee, Yong Jae},
  booktitle={Proceedings of the IEEE/CVF conference on computer vision and pattern recognition},
  pages={26296--26306},
  year={2024}
}

@misc{qwen,
      title={Qwen2.5-VL Technical Report}, 
      author={Shuai Bai and Keqin Chen and Xuejing Liu and Jialin Wang and Wenbin Ge and Sibo Song and Kai Dang and Peng Wang and Shijie Wang and Jun Tang and Humen Zhong and Yuanzhi Zhu and Mingkun Yang and Zhaohai Li and Jianqiang Wan and Pengfei Wang and Wei Ding and Zheren Fu and Yiheng Xu and Jiabo Ye and Xi Zhang and Tianbao Xie and Zesen Cheng and Hang Zhang and Zhibo Yang and Haiyang Xu and Junyang Lin},
      journal={arXiv preprint arXiv:2502.13923},
      year={2025}
}

@article{llava_video,
  title={Llava-video: Video instruction tuning with synthetic data},
  author={Zhang, Yuanhan and Wu, Jinming and Li, Wei and Li, Bo and Ma, Zejun and Liu, Ziwei and Li, Chunyuan},
  journal={arXiv preprint arXiv:2410.02713},
  year={2024}
}

@inproceedings{fastv,
  title={An image is worth 1/2 tokens after layer 2: Plug-and-play inference acceleration for large vision-language models},
  author={Chen, Liang and Zhao, Haozhe and Liu, Tianyu and Bai, Shuai and Lin, Junyang and Zhou, Chang and Chang, Baobao},
  booktitle={European Conference on Computer Vision},
  pages={19--35},
  year={2024},
  organization={Springer}
}

@article{pyramiddrop,
  title={Pyramiddrop: Accelerating your large vision-language models via pyramid visual redundancy reduction},
  author={Xing, Long and Huang, Qidong and Dong, Xiaoyi and Lu, Jiajie and Zhang, Pan and Zang, Yuhang and Cao, Yuhang and He, Conghui and Wang, Jiaqi and Wu, Feng and others},
  journal={arXiv preprint arXiv:2410.17247},
  year={2024}
}

@article{sparsevlm,
  title={Sparsevlm: Visual token sparsification for efficient vision-language model inference},
  author={Zhang, Yuan and Fan, Chun-Kai and Ma, Junpeng and Zheng, Wenzhao and Huang, Tao and Cheng, Kuan and Gudovskiy, Denis and Okuno, Tomoyuki and Nakata, Yohei and Keutzer, Kurt and others},
  journal={arXiv preprint arXiv:2410.04417},
  year={2024}
}

@inproceedings{prumerge,
  title={Llava-prumerge: Adaptive token reduction for efficient large multimodal models},
  author={Shang, Yuzhang and Cai, Mu and Xu, Bingxin and Lee, Yong Jae and Yan, Yan},
  booktitle={Proceedings of the IEEE/CVF International Conference on Computer Vision},
  pages={22857--22867},
  year={2025}
}

@inproceedings{visionzip,
  title={Visionzip: Longer is better but not necessary in vision language models},
  author={Yang, Senqiao and Chen, Yukang and Tian, Zhuotao and Wang, Chengyao and Li, Jingyao and Yu, Bei and Jia, Jiaya},
  booktitle={Proceedings of the IEEE/CVF Conference on Computer Vision and Pattern Recognition},
  pages={19792--19802},
  year={2025}
}

@inproceedings{divprune,
  title={Divprune: Diversity-based visual token pruning for large multimodal models},
  author={Alvar, Saeed Ranjbar and Singh, Gursimran and Akbari, Mohammad and Zhang, Yong},
  booktitle={Proceedings of the Computer Vision and Pattern Recognition Conference},
  pages={9392--9401},
  year={2025}
}

@article{cdpruner,
  title={Beyond attention or similarity: Maximizing conditional diversity for token pruning in mllms},
  author={Zhang, Qizhe and Liu, Mengzhen and Li, Lichen and Lu, Ming and Zhang, Yuan and Pan, Junwen and She, Qi and Zhang, Shanghang},
  journal={Advances in Neural Information Processing Systems},
  volume={38},
  pages={25438--25468},
  year={2025}
}

@article{eclip,
  title={Exploring visual interpretability for contrastive language-image pre-training},
  author={Li, Yi and Wang, Hualiang and Duan, Yiqun and Xu, Hang and Li, Xiaomeng},
  journal={arXiv preprint arXiv:2209.07046},
  year={2022}
}

@inproceedings{lmms_eval,
  title={Lmms-eval: Reality check on the evaluation of large multimodal models},
  author={Zhang, Kaichen and Li, Bo and Zhang, Peiyuan and Pu, Fanyi and Cahyono, Joshua Adrian and Hu, Kairui and Liu, Shuai and Zhang, Yuanhan and Yang, Jingkang and Li, Chunyuan and others},
  booktitle={Findings of the Association for Computational Linguistics: NAACL 2025},
  pages={881--916},
  year={2025}
}

@inproceedings{vqav2,
  title={Making the v in vqa matter: Elevating the role of image understanding in visual question answering},
  author={Goyal, Yash and Khot, Tejas and Summers-Stay, Douglas and Batra, Dhruv and Parikh, Devi},
  booktitle={Proceedings of the IEEE conference on computer vision and pattern recognition},
  pages={6904--6913},
  year={2017}
}

@inproceedings{textvqa,
  title={Towards vqa models that can read},
  author={Singh, Amanpreet and Natarajan, Vivek and Shah, Meet and Jiang, Yu and Chen, Xinlei and Batra, Dhruv and Parikh, Devi and Rohrbach, Marcus},
  booktitle={Proceedings of the IEEE/CVF conference on computer vision and pattern recognition},
  pages={8317--8326},
  year={2019}
}

@inproceedings{mme,
  title     = {MME: A Comprehensive Evaluation Benchmark for Multimodal Large Language Models},
  author    = {Fu, Chaoyou and Chen, Peixian and Shen, Yunhang and Qin, Yulei and Zhang, Mengdan and Lin, Xu and Yang, Jinrui and Zheng, Xiawu and Li, Ke and Sun, Xing and Wu, Yunsheng and Ji, Rongrong and Shan, Caifeng and He, Ran},
  booktitle = {Advances in Neural Information Processing Systems},
  volume    = {38},
  year      = {2025}
}

@inproceedings{pope,
  title={Evaluating object hallucination in large vision-language models},
  author={Li, Yifan and Du, Yifan and Zhou, Kun and Wang, Jinpeng and Zhao, Xin and Wen, Ji-Rong},
  booktitle={Proceedings of the 2023 conference on empirical methods in natural language processing},
  pages={292--305},
  year={2023}
}

@inproceedings{mmbench,
  title={Mmbench: Is your multi-modal model an all-around player?},
  author={Liu, Yuan and Duan, Haodong and Zhang, Yuanhan and Li, Bo and Zhang, Songyang and Zhao, Wangbo and Yuan, Yike and Wang, Jiaqi and He, Conghui and Liu, Ziwei and others},
  booktitle={European conference on computer vision},
  pages={216--233},
  year={2024},
  organization={Springer}
}

@inproceedings{video_mme,
  title={Video-mme: The first-ever comprehensive evaluation benchmark of multi-modal llms in video analysis},
  author={Fu, Chaoyou and Dai, Yuhan and Luo, Yongdong and Li, Lei and Ren, Shuhuai and Zhang, Renrui and Wang, Zihan and Zhou, Chenyu and Shen, Yunhang and Zhang, Mengdan and others},
  booktitle={Proceedings of the IEEE/CVF conference on computer vision and pattern recognition},
  pages={24108--24118},
  year={2025}
}

@article{egoschema,
  title={Egoschema: A diagnostic benchmark for very long-form video language understanding},
  author={Mangalam, Karttikeya and Akshulakov, Raiymbek and Malik, Jitendra},
  journal={Advances in Neural Information Processing Systems},
  volume={36},
  pages={46212--46244},
  year={2023}
}

@inproceedings{tempcompass,
  title={Tempcompass: Do video llms really understand videos?},
  author={Liu, Yuanxin and Li, Shicheng and Liu, Yi and Wang, Yuxiang and Ren, Shuhuai and Li, Lei and Chen, Sishuo and Sun, Xu and Hou, Lu},
  booktitle={Findings of the Association for Computational Linguistics: ACL 2024},
  pages={8731--8772},
  year={2024}
}

@inproceedings{gqa,
  title={Gqa: A new dataset for real-world visual reasoning and compositional question answering},
  author={Hudson, Drew A and Manning, Christopher D},
  booktitle={Proceedings of the IEEE/CVF conference on computer vision and pattern recognition},
  pages={6700--6709},
  year={2019}
}

@article{scienceqa,
  title={Learn to explain: Multimodal reasoning via thought chains for science question answering},
  author={Lu, Pan and Mishra, Swaroop and Xia, Tanglin and Qiu, Liang and Chang, Kai-Wei and Zhu, Song-Chun and Tafjord, Oyvind and Clark, Peter and Kalyan, Ashwin},
  journal={Advances in neural information processing systems},
  volume={35},
  pages={2507--2521},
  year={2022}
}

@article{mmvet,
  title={Mm-vet: Evaluating large multimodal models for integrated capabilities},
  author={Yu, Weihao and Yang, Zhengyuan and Li, Linjie and Wang, Jianfeng and Lin, Kevin and Liu, Zicheng and Wang, Xinchao and Wang, Lijuan},
  journal={arXiv preprint arXiv:2308.02490},
  year={2023}
}

@inproceedings{docvqa,
  title={Docvqa: A dataset for vqa on document images},
  author={Mathew, Minesh and Karatzas, Dimosthenis and Jawahar, CV},
  booktitle={Proceedings of the IEEE/CVF winter conference on applications of computer vision},
  pages={2200--2209},
  year={2021}
}

@inproceedings{ai2d,
  title={A diagram is worth a dozen images},
  author={Kembhavi, Aniruddha and Salvato, Mike and Kolve, Eric and Seo, Minjoon and Hajishirzi, Hannaneh and Farhadi, Ali},
  booktitle={European conference on computer vision},
  pages={235--251},
  year={2016},
  organization={Springer}
}

@inproceedings{mmmu,
  title={Mmmu: A massive multi-discipline multimodal understanding and reasoning benchmark for expert agi},
  author={Yue, Xiang and Ni, Yuansheng and Zhang, Kai and Zheng, Tianyu and Liu, Ruoqi and Zhang, Ge and Stevens, Samuel and Jiang, Dongfu and Ren, Weiming and Sun, Yuxuan and others},
  booktitle={Proceedings of the IEEE/CVF conference on computer vision and pattern recognition},
  pages={9556--9567},
  year={2024}
}

@inproceedings{clip,
  title={Learning transferable visual models from natural language supervision},
  author={Radford, Alec and Kim, Jong Wook and Hallacy, Chris and Ramesh, Aditya and Goh, Gabriel and Agarwal, Sandhini and Sastry, Girish and Askell, Amanda and Mishkin, Pamela and Clark, Jack and others},
  booktitle={International conference on machine learning},
  pages={8748--8763},
  year={2021},
  organization={PmLR}
}

@inproceedings{chartqa,
  title={Chartqa: A benchmark for question answering about charts with visual and logical reasoning},
  author={Masry, Ahmed and Do, Xuan Long and Tan, Jia Qing and Joty, Shafiq and Hoque, Enamul},
  booktitle={Findings of the association for computational linguistics: ACL 2022},
  pages={2263--2279},
  year={2022}
}

@misc{llavanext,
    title={LLaVA-NeXT: Improved reasoning, OCR, and world knowledge},
    url={https://llava-vl.github.io/blog/2024-01-30-llava-next/},
    author={Liu, Haotian and Li, Chunyuan and Li, Yuheng and Li, Bo and Zhang, Yuanhan and Shen, Sheng and Lee, Yong Jae},
    month={January},
    year={2024}
}

@article{ocrbench,
  title={Ocrbench: on the hidden mystery of ocr in large multimodal models},
  author={Liu, Yuliang and Li, Zhang and Huang, Mingxin and Yang, Biao and Yu, Wenwen and Li, Chunyuan and Yin, Xu-Cheng and Liu, Cheng-Lin and Jin, Lianwen and Bai, Xiang},
  journal={Science China Information Sciences},
  volume={67},
  number={12},
  pages={220102},
  year={2024},
  publisher={Springer}
}
\newpage
\title{Supplementary Material\\
AnchorPrune: Relevance-Anchored Contextual Expansion for Visual Token Pruning}

\titlerunning{Supplementary Material for AnchorPrune}

\author{Kyuan Oh\orcidlink{0009-0001-8047-6396} \and
Bumsoo Kim\thanks{: corresponding author}\orcidlink{0009-0003-6519-3253}}

\authorrunning{K.~Oh and B.~Kim}

\institute{Chung-Ang University, Seoul, Korea\\
\email{\{oka04108,bumsoo\}@cau.ac.kr}}

\maketitle

\appendix

\setcounter{table}{0}
\setcounter{figure}{0}
\setcounter{equation}{0}

\renewcommand{\thetable}{S\arabic{table}}
\renewcommand{\thefigure}{S\arabic{figure}}
\renewcommand{\theequation}{S\arabic{equation}}

\renewcommand{\theHsection}{supp.\Alph{section}}
\renewcommand{\theHsubsection}{supp.\Alph{section}.\arabic{subsection}}
\renewcommand{\theHtable}{supp.table.\arabic{table}}
\renewcommand{\theHfigure}{supp.figure.\arabic{figure}}
\renewcommand{\theHequation}{supp.equation.\arabic{equation}}

\vspace{0.5em}

\section{Implementation Details and Experimental Protocol}
\label{sec:supp_implementation}

This section specifies the implementation choices and experimental protocol used throughout our evaluation. We describe the model-specific candidate-token sequences, pruning locations, architecture-dependent signals, adaptive anchor-budget configurations, controlled comparison protocol, and hardware and efficiency-measurement settings.

\subsection{Model-Specific Visual Tokenization}
\label{sec:supp_models}

Table~\ref{tab:model_tokenization} summarizes the candidate visual-token sequence processed by AnchorPrune for each backbone. To ensure a controlled comparison, all pruning methods evaluated on the same backbone use an identical fixed input resolution, visual preprocessing pipeline, tokenization procedure, and candidate sequence. This prevents changes in resolution or token granularity from confounding the effect of the pruning method itself. 

\begin{table}[h]
\centering
\caption{\textbf{Model-specific input and visual-token configurations.} The tokenization structure denotes the candidate sequence on which visual-token selection is performed.}
\label{tab:model_tokenization}
\setlength{\tabcolsep}{4.0pt}
\renewcommand{\arraystretch}{1.05}
\resizebox{\columnwidth}{!}{%
\begin{tabular}{lccc}
\toprule
Model & Input Configuration & Tokenization Structure & Visual Tokens \\
\midrule
LLaVA-1.5-7B & \(336\times336\) image & \(24\times24\) grid & 576 \\
LLaVA-NeXT-7B & \(672\times672\) image & \(5\times24\times24\) grids & 2,880 \\
Qwen2.5-VL-7B & \(1008\times1008\) image & \(36\times36\) merged grid & 1,296 \\
LLaVA-Video-7B & 16 frames at \(384\times384\) & \(16\times13\times13\) post-pooling grid & 2,704 \\
\bottomrule
\end{tabular}}
\end{table}

For LLaVA-NeXT-7B~\cite{llavanext}, the five visual units form the model's high-resolution multi-crop representation, with each unit contributing a \(24\times24\) spatial grid. For LLaVA-Video-7B~\cite{llava_video}, we uniformly sample 16 frames, process each frame at \(384\times384\), and represent it using a \(13\times13\) post-pooling grid before concatenating the frame-level sequences in temporal order.

\subsection{Pruning Location and Native Token Ordering}
\label{sec:supp_insertion}

AnchorPrune is applied after visual feature extraction and before the retained visual sequence is consumed by the language model. It leaves the parameters and layer structure of the vision encoder, multimodal projector, and language model unchanged.

For LLaVA-1.5-7B and LLaVA-NeXT-7B, pruning operates on the visual tokens associated with the input image or each high-resolution visual unit. For Qwen2.5-VL-7B, selection is applied after the model-specific spatial processing and token merging that produce the visual sequence supplied to the language model. For LLaVA-Video-7B, post-pooling tokens from the 16 sampled frames are concatenated into a frame-major sequence before selection.

After pruning, retained tokens are restored to their native order before language-model inference. Image models preserve raster-scan order within each visual unit and the original ordering across multiple units, whereas LLaVA-Video-7B preserves temporal frame order first and raster-scan order within each frame. This operation changes only the sequence order of the selected tokens and does not alter subset membership. Section~\ref{sec:supp_sequence_order} isolates its effect on spatially sensitive benchmarks.

\subsection{Architecture-Specific Signal Extraction}
\label{sec:supp_architecture}

The main paper defines the anchoring-priority score, novelty measure, and global-importance prior. This section details the architecture-specific representation spaces and attention signals used to instantiate these quantities.

\subsubsection*{LLaVA-1.5-7B and LLaVA-NeXT-7B.} For LLaVA-1.5-7B~\cite{llava} and LLaVA-NeXT-7B~\cite{llavanext}, Stage~1 matching and novelty use normalized pre-projector visual features in the paired CLIP representation space~\cite{clip}, with instruction embeddings produced by the corresponding CLIP text encoder. Instructions that exceed the supported context length are partitioned into valid segments, and the resulting patch--text scores are averaged across segments. The Stage~2 global-importance prior is obtained from a late vision-transformer layer by averaging \texttt{[CLS]}-to-patch attention across heads. Section~\ref{sec:clip_direction} compares raw and negated CLIP patch--text similarity.

\subsubsection*{Qwen2.5-VL-7B.} Because Qwen2.5-VL-7B~\cite{qwen} does not expose a paired pre-projector vision--language space, Stage~1 matching is performed after the multimodal projector. Projected visual tokens and instruction-token embeddings are normalized, and each visual token is assigned its maximum cosine similarity over instruction tokens. This preserves alignment with individual semantic components that may be diluted by averaging the instruction into a single representation. Section~\ref{sec:supp_qwen_matching} compares maximum matching with mean aggregation.

Qwen2.5-VL-7B spatially merges vision-encoder tokens before constructing the final candidate sequence, so the attention-derived importance prior is aligned with the post-merge candidates. We first average the attention mass received by each pre-merge token across source positions and heads, and then average these values within the pre-merge group associated with each final candidate. Projected candidate features are used for both adaptive-anchor novelty and Stage~2 contextual novelty.

\subsubsection*{LLaVA-Video-7B.} For LLaVA-Video-7B~\cite{llava_video}, the CLIP-aligned implementation is applied independently to each sampled frame, after which frame-level candidates are concatenated into a temporally ordered sequence. Selection can therefore remove both spatial redundancy within individual frames and repeated evidence across adjacent frames. The retained sequence is restored to frame-major temporal order before language-model inference.

\subsection{Adaptive Anchor-Budget Configuration}
\label{sec:supp_hyperparameters}

We use a shared novelty threshold \(\tau=0.2\) and patience parameter \(P=3\) across all models and benchmarks. For a visual unit with assigned retained-token budget \(K_u\), the maximum anchor size is \(K_{\max,u}=\lfloor K_u/2\rfloor\). For single-unit image models, \(K_u=K\); for LLaVA-NeXT-7B and LLaVA-Video-7B, \(K_u\) denotes the budget assigned to each image unit or frame. Neither \(\tau\), \(P\), nor \(K_{\min}\) is tuned per benchmark. Table~\ref{tab:hyperparameters} reports the complete \(K_{\min}\) schedule. A single configuration is used for every benchmark evaluated at a given backbone and retained-token budget.

\begin{table}[h]
\centering
\caption{\textbf{Adaptive anchor-budget configurations.} The novelty threshold and patience are fixed to \(\tau=0.2\) and \(P=3\) across all experiments. The minimum anchor size \(K_{\min}\) follows a fixed budget-dependent schedule and is not tuned per benchmark.}
\label{tab:hyperparameters}
\setlength{\tabcolsep}{5pt}
\renewcommand{\arraystretch}{1.05}
\resizebox{\columnwidth}{!}{%
\begin{tabular}{llccc}
\toprule
Model / Setting & Unit & \#Units & Total Budget \(K\) & \(K_{\min}\) per Unit \\
\midrule
\multirow{3}{*}{LLaVA-1.5-7B} & image & 1 & 32 & 5 \\
& image & 1 & 64 & 10 \\
& image & 1 & 128 & 20 \\
\midrule
\multirow{3}{*}{Qwen2.5-VL-7B} & image & 1 & 64 & 10 \\
& image & 1 & 128 & 20 \\
& image & 1 & 256 & 40 \\
\midrule
\multirow{3}{*}{LLaVA-NeXT-7B} & image unit & 5 & 160 & 5 \\
& image unit & 5 & 320 & 10 \\
& image unit & 5 & 640 & 20 \\
\midrule
\multirow{2}{*}{LLaVA-Video-7B} & frame & 16 & 512 & 5 \\
& frame & 16 & 1024 & 10 \\
\bottomrule
\end{tabular}}
\end{table}

For LLaVA-NeXT-7B and LLaVA-Video-7B, the protected relevance anchor is constructed independently within each visual unit or frame using the corresponding per-unit \(K_{\min}\) schedule and \(K_{\max,u}\). The resulting unit-level anchors are concatenated to form the global protected anchor, after which the remaining total budget is allocated globally by Stage~2 over the full candidate sequence. Retained tokens are finally restored to their native unit/frame-major order.

\subsubsection*{Fallback behavior.} Within each visual unit, the adaptive procedure terminates before \(K_{\max,u}\) only when the relevance-ranked sequence contains the required \(P\) novelty events. Otherwise, it conservatively uses \(K_{\max,u}\). Because each unit-level anchor is bounded by half of its assigned budget, at least half of the total retained-token budget remains available for global contextual expansion.

\subsection{Evaluation Protocol}
\label{sec:supp_evaluation}

For each backbone and retained-token budget, all methods use the same checkpoint, input resolution, visual preprocessing, tokenization procedure, prompt template, decoding configuration, benchmark split, and evaluation pipeline. Only the visual-token selection rule is changed. Fixing the complete preprocessing and tokenization pipeline is essential because changes in resolution or spatial merging alter both the number and granularity of candidate tokens, thereby confounding the effect of the pruning method itself.

We use \texttt{lmms-eval}~\cite{lmms_eval} whenever the corresponding model--benchmark pair is supported and otherwise follow the official benchmark evaluator. Baseline implementations follow their published selection rules and recommended configurations. A method is evaluated on a backbone only when its required architectural signals and candidate-token structure are available without modifying its underlying selection objective. This accounts for the smaller comparison sets reported for Qwen2.5-VL-7B and LLaVA-Video-7B. All comparisons use exactly matched final retained-token budgets.

Benchmark definitions and the computation of \emph{Rel.} are provided in the main paper. In the detailed video evaluation, Video-MME~\cite{video_mme} with and without subtitles are treated as separate aggregate metrics, EgoSchema~\cite{egoschema} uses the official 500-example subset, and TempCompass~\cite{tempcompass} contributes its average score.

\subsection{Hardware and Efficiency Measurement}
\label{sec:supp_hardware}

Image experiments on LLaVA-1.5-7B, LLaVA-NeXT-7B, and Qwen2.5-VL-7B use four NVIDIA RTX A5000 GPUs, whereas video experiments on LLaVA-Video-7B use two NVIDIA RTX 4090 GPUs. Within each reported comparison, all methods are evaluated in the same hardware and software environment.

The detailed efficiency experiment in Sec.~\ref{sec:supp_efficiency} uses LLaVA-1.5-7B with 32 retained visual tokens. Prefill latency includes visual-token scoring and selection overhead in addition to the reduced model forward pass, while decode latency is reported per generated token. Estimated FLOPs, resident GPU memory, and MME performance are measured under the same inference setting.

\section{Detailed Video Results}
\label{sec:supp_video_results}

The main paper reports aggregate video results. Table~\ref{tab:video_detailed} provides the complete Video-MME~\cite{video_mme} breakdown by duration and subtitle condition, together with EgoSchema~\cite{egoschema} and TempCompass~\cite{tempcompass}. We compare AnchorPrune against FastV~\cite{fastv}, DivPrune~\cite{divprune}, and CDPruner~\cite{cdpruner}.

\begin{table}[H]
\centering
\caption{\textbf{Detailed video results on LLaVA-Video-7B under 16-frame evaluation.} Video-MME is decomposed into short-, medium-, and long-duration subsets with and without subtitles. EgoSchema uses the 500-example subset, and TempCompass reports its average score. Best and second-best \emph{Rel.} values within each budget are shown in \textbf{bold} and \underline{underlined}, respectively.}
\label{tab:video_detailed}
\setlength{\tabcolsep}{3.3pt}
\renewcommand{\arraystretch}{1.05}
\resizebox{\textwidth}{!}{%
\begin{tabular}{lcccc cccc cc|c}
\toprule
\multirow{2}{*}{Method} &
\multicolumn{4}{c}{Video-MME w/o Subtitles} &
\multicolumn{4}{c}{Video-MME w/ Subtitles} &
EgoSchema & TempCompass & \multirow{2}{*}{Rel.} \\
\cmidrule(lr){2-5}\cmidrule(lr){6-9}
& Short & Medium & Long & Total
& Short & Medium & Long & Total
& 500-subset & Avg. & \\
\midrule
\multicolumn{12}{c}{\textbf{Upper Bound: All 2,704 Visual Tokens}} \\
\midrule
LLaVA-Video-7B & 70.78 & 58.67 & 50.44 & 59.96 & 72.78 & 59.78 & 51.44 & 61.33 & 55.4 & 67.1 & 100.0 \\
\midrule
\multicolumn{12}{c}{\textbf{Retain 1,024 Visual Tokens} ($\downarrow$ 62.1\%)} \\
\midrule
FastV (ECCV'24)       & 69.70 & 57.70 & 50.40 & 59.26 & 71.20 & 58.70 & 51.90 & 60.59 & 52.8 & 65.6 & \second{97.7} \\
DivPrune (CVPR'25)    & 69.11 & 54.44 & 47.44 & 57.00 & 70.78 & 56.33 & 50.33 & 59.15 & 53.2 & 64.9 & 96.1 \\
CDPruner (NeurIPS'25) & 69.11 & 55.56 & 48.89 & 57.85 & 70.89 & 57.33 & 51.22 & 59.81 & 52.6 & 65.6 & 96.8 \\
\ourrow \ours          & 69.56 & 55.89 & 49.11 & 58.19 & 72.78 & 58.56 & 51.33 & 60.89 & 55.0 & 65.4 & \best{98.3} \\
\midrule
\multicolumn{12}{c}{\textbf{Retain 512 Visual Tokens} ($\downarrow$ 81.1\%)} \\
\midrule
FastV (ECCV'24)       & 65.40 & 55.80 & 47.90 & 56.37 & 66.80 & 56.10 & 50.00 & 57.63 & 50.2 & 61.1 & 92.4 \\
DivPrune (CVPR'25)    & 67.44 & 52.56 & 47.67 & 55.89 & 68.78 & 56.33 & 49.44 & 58.19 & 52.0 & 61.3 & \second{93.4} \\
CDPruner (NeurIPS'25) & 65.33 & 52.00 & 45.33 & 54.22 & 65.89 & 55.78 & 49.33 & 57.00 & 51.4 & 60.6 & 91.7 \\
\ourrow \ours          & 67.22 & 54.22 & 47.33 & 56.26 & 69.56 & 54.89 & 49.78 & 58.07 & 53.0 & 61.8 & \best{94.1} \\
\bottomrule
\end{tabular}}
\end{table}

At 1,024 retained tokens, AnchorPrune achieves the highest aggregate retention, preserving 98.3\% of full-token performance. Its advantage is especially pronounced on EgoSchema, where it obtains 55.0 compared with 52.6--53.2 for the competing methods. At 512 tokens, AnchorPrune again yields the highest \emph{Rel.} at 94.1\%, together with the strongest EgoSchema and TempCompass scores among the pruning methods.

No method dominates every Video-MME duration and subtitle condition. AnchorPrune nevertheless provides the strongest overall balance across short-, medium-, and long-duration videos and across the complementary video benchmarks. This pattern supports relevance-anchored contextual expansion for temporally structured inputs, where repeated evidence across adjacent frames creates substantial redundancy.

\section{Additional Ablation Studies}
\label{sec:supp_ablations}

\subsection{Effect of Retained-Token Ordering}
\label{sec:supp_sequence_order}

AnchorPrune restores retained tokens to their native spatial or temporal order before language-model inference. Table~\ref{tab:sequence_order} isolates this operation by evaluating an identical selected subset under raw selection order and native raster-scan order on MME-OCR~\cite{mme}, OCRBench~\cite{ocrbench}, and TextVQA~\cite{textvqa}.

\begin{table}[H]
\centering
\caption{\textbf{Effect of retained-token ordering on spatially sensitive benchmarks.} Both variants use exactly the same selected subset and differ only in the sequence order supplied to the language model.}
\label{tab:sequence_order}
\setlength{\tabcolsep}{8pt}
\renewcommand{\arraystretch}{1.05}
\begin{tabular}{lccc}
\toprule
Sequence Order & MME-OCR & OCRBench & TextVQA \\
\midrule
\multicolumn{4}{c}{\textbf{Retain 64 Tokens}} \\
\midrule
Raw selection order & 132.5 & 26.9 & 55.5 \\
\ourrow Raster-scan order & \best{140.0} & \best{28.9} & \best{56.1} \\
\bottomrule
\end{tabular}
\end{table}

Restoring raster-scan order improves all three benchmarks, with the largest gains on MME-OCR and OCRBench, which are especially sensitive to spatial organization and local reading order. Because subset membership is held fixed, the improvement is attributable solely to presenting the retained evidence in the positional progression expected by the pretrained model.

\subsection{Sensitivity to Adaptive Anchor-Budget Hyperparameters}
\label{sec:supp_sensitivity}

Table~\ref{tab:adaptive_sensitivity} evaluates sensitivity to the novelty threshold \(\tau\) and patience parameter \(P\) on MME~\cite{mme}. Each experiment varies only the parameter under study while holding all remaining settings fixed.

\begin{table*}[h]
\centering
\caption{\textbf{Sensitivity of the adaptive anchor budget on MME.}
Results are reported under the most aggressive retained-token budget for each backbone: 32 of 576 tokens for LLaVA-1.5-7B, 160 of 2,880 tokens for LLaVA-NeXT-7B, and 64 of 1,296 tokens for Qwen2.5-VL-7B.
Left: novelty threshold \(\tau\) with \(P=3\).
Right: patience parameter \(P\) with \(\tau=0.2\).
All remaining settings are fixed.}
\label{tab:adaptive_sensitivity}
\setlength{\tabcolsep}{4.2pt}
\renewcommand{\arraystretch}{1.02}
\small

\begin{minipage}[t]{0.485\textwidth}
    \centering
    \textbf{Novelty threshold \(\tau\)}\\[3pt]
    \begin{tabular}{cccc}
    \toprule
    \(\tau\) &
    \shortstack{LLaVA-\\1.5-7B} &
    \shortstack{LLaVA-\\NeXT-7B} &
    \shortstack{Qwen2.5-\\VL-7B} \\
    \midrule
    0.1 & 1392.2 & 1469.3 & 1560.2 \\
    \ourrow 0.2 & \best{1394.7} & 1481.1 & \best{1563.2} \\
    0.3 & 1388.9 & \best{1489.1} & 1562.8 \\
    0.4 & 1390.6 & 1480.3 & 1557.3 \\
    \bottomrule
    \end{tabular}
\end{minipage}
\hfill
\begin{minipage}[t]{0.485\textwidth}
    \centering
    \textbf{Patience parameter \(P\)}\\[3pt]
    \begin{tabular}{cccc}
    \toprule
    \(P\) &
    \shortstack{LLaVA-\\1.5-7B} &
    \shortstack{LLaVA-\\NeXT-7B} &
    \shortstack{Qwen2.5-\\VL-7B} \\
    \midrule
    1 & 1384.8 & 1480.2 & 1559.2 \\
    \ourrow 3 & \best{1394.7} & \best{1481.1} & \best{1563.2} \\
    5 & 1386.2 & 1478.6 & 1552.2 \\
    \bottomrule
    \end{tabular}
\end{minipage}
\end{table*}

Performance is stable across the tested threshold range. The shared setting \(\tau=0.2\) is best on LLaVA-1.5-7B and Qwen2.5-VL-7B, while \(\tau=0.3\) is moderately better on LLaVA-NeXT-7B. We retain \(\tau=0.2\) as a single architecture-independent setting rather than selecting a backbone-specific optimum.

The patience study shows a clearer pattern: \(P=3\) achieves the highest MME score on all three architectures. A smaller value can terminate anchor construction before sufficiently distributed evidence is retained, whereas a larger value can delay the transition to contextual expansion. These results support \(P=3\) as a stable shared configuration.

\subsection{Stage-1 CLIP Similarity Direction}
\label{sec:clip_direction}

For CLIP-aligned VLMs~\cite{clip}, AnchorPrune uses negated patch--text
similarity as the Stage~1 anchoring-priority signal. This choice is motivated
by prior analysis of CLIP's dense image--text similarity maps, which found
that raw local similarities can respond more strongly to background regions
than to discriminative foreground content and that reversing the similarity
map can improve localization~\cite{eclip}.

\begin{figure*}[h]
    \centering
    \includegraphics[width=\textwidth]
    {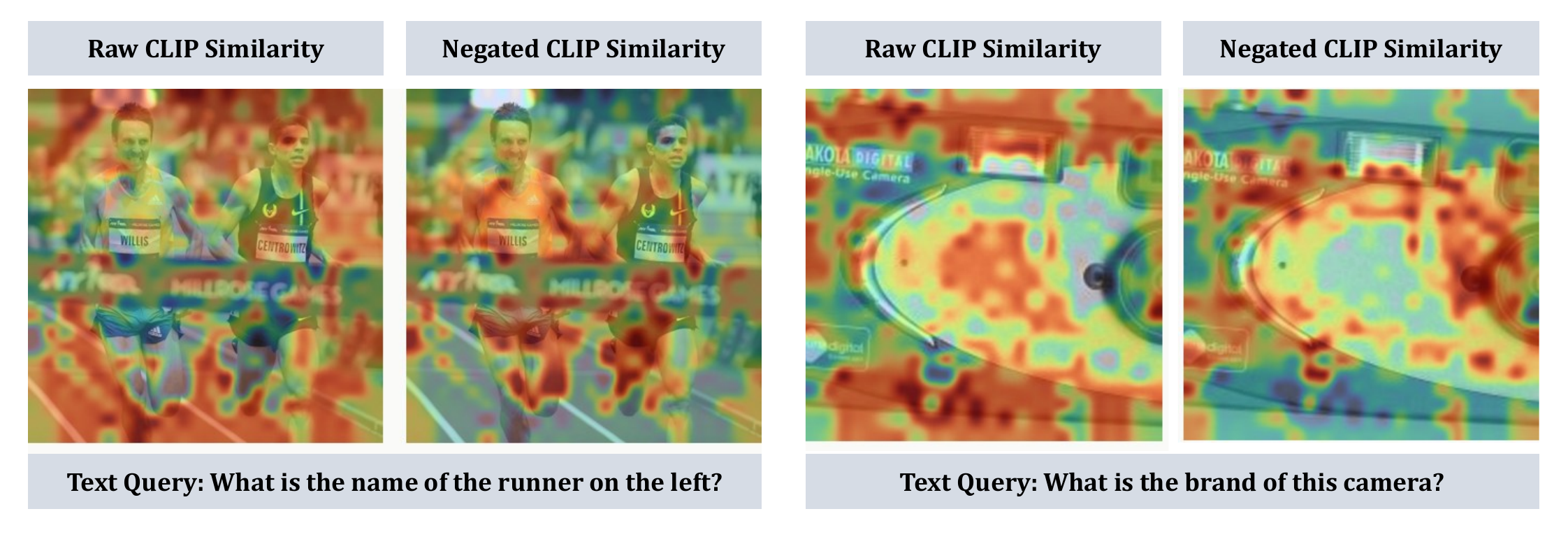}
    \caption{\textbf{Qualitative comparison of raw and negated CLIP
    patch--text similarity.} Raw similarity can emphasize broad background
    or non-discriminative regions, whereas negated similarity assigns greater
    priority to localized regions associated with the queried evidence in these examples,
    including instruction-relevant objects and OCR-relevant regions. The heatmaps
    visualize relative Stage~1 priority rather than calibrated semantic
    relevance.}
    \label{fig:clip_similarity_direction}
\end{figure*}

Figure~\ref{fig:clip_similarity_direction} visualizes the spatial priority
induced by raw and negated similarity. Raw similarity can emphasize broad
background or non-discriminative regions, whereas negated similarity assigns
greater priority to localized regions associated with the queried evidence in these examples.

Table~\ref{tab:clip_direction} evaluates the downstream effect of the
similarity direction under identical adaptive anchor-budget and Stage~2
contextual-expansion settings. Negated similarity consistently improves MME and
POPE at both retained-token budgets, while TextVQA remains comparatively
stable. At 32 retained tokens, it improves MME from 1245.4 to 1394.7 and
POPE from 74.2 to 85.3. These results indicate that the gain is not attributable
to a uniform rescaling of token scores; rather, the negated direction induces
a Stage~1 ranking that better preserves evidence required for broad visual
perception and hallucination-sensitive evaluation.

Consistent with ECLIP~\cite{eclip}, we use negated similarity strictly as an
empirical anchoring-priority signal for the evaluated CLIP-aligned models.
We do not interpret it as calibrated semantic relevance or claim that raw
CLIP similarity is universally inverted. The qualitative and quantitative
results instead indicate that the negated direction is better suited to
constructing the protected Stage~1 relevance anchor.

\begin{table}[t]
\centering
\caption{\textbf{Effect of CLIP similarity direction on Stage~1 anchoring.}
Raw and negated CLIP patch--text similarity are compared on LLaVA-1.5-7B
under identical adaptive anchor-budget and Stage~2 contextual-expansion settings.}
\label{tab:clip_direction}
\setlength{\tabcolsep}{7pt}
\renewcommand{\arraystretch}{1.05}
\begin{tabular}{lccc}
\toprule
Stage~1 CLIP Score & MME & TextVQA & POPE \\
\midrule
\multicolumn{4}{c}{\textbf{Upper Bound: All 576 Tokens}} \\
\midrule
Full visual tokens & 1509.1 & 58.3 & 85.8 \\
\midrule
\multicolumn{4}{c}{\textbf{Retain 64 Tokens}} \\
\midrule
Raw CLIP similarity & 1355.8 & 56.0 & 80.2 \\
\ourrow Negated CLIP similarity &
\best{1420.2} & \best{56.1} & \best{85.8} \\
\midrule
\multicolumn{4}{c}{\textbf{Retain 32 Tokens}} \\
\midrule
Raw CLIP similarity & 1245.4 & 53.7 & 74.2 \\
\ourrow Negated CLIP similarity &
\best{1394.7} & \best{54.2} & \best{85.3} \\
\bottomrule
\end{tabular}
\end{table}

\subsection{Post-Projector Relevance Matching}
\label{sec:supp_qwen_matching}

For Qwen2.5-VL-7B~\cite{qwen}, Stage~1 matching is computed in the post-projector language-model input space. Table~\ref{tab:qwen_matching} compares token-wise maximum matching with mean aggregation of the instruction-token embeddings.

\begin{table}[h]
\centering
\caption{\textbf{Post-projector relevance matching in Qwen2.5-VL-7B.} Mean aggregation and token-wise maximum matching are compared under the same retained budget of 256 visual tokens.}
\label{tab:qwen_matching}
\setlength{\tabcolsep}{4.2pt}
\renewcommand{\arraystretch}{1.05}
\resizebox{\columnwidth}{!}{%
\begin{tabular}{lccccccc}
\toprule
Relevance Formulation & MME & TextVQA & DocVQA & AI2D & MMMU & MMB$^{\mathrm{EN}}$ & MMB$^{\mathrm{CN}}$ \\
\midrule
Mean aggregation & 1647.4 & 72.0 & 72.2 & 78.7 & 47.7 & 80.2 & 77.3 \\
\ourrow Token-wise max matching & \best{1657.5} & \best{72.6} & \best{72.9} & \best{78.8} & \best{48.6} & \best{80.8} & \best{78.2} \\
\bottomrule
\end{tabular}}
\end{table}

Token-wise maximum matching improves all seven benchmarks. Although the individual gains are moderate, their consistency indicates that averaging instruction-token embeddings can obscure alignment with specific semantic components. Maximum matching preserves each visual token's strongest correspondence to any instruction token and is therefore better suited to Stage~1 ranking in the post-projector representation space.

\section{Detailed Efficiency Analysis}
\label{sec:supp_efficiency}

Table~\ref{tab:efficiency} reports the raw measurements underlying the efficiency--performance comparison in the main paper. We compare against FastV~\cite{fastv}, PyramidDrop~\cite{pyramiddrop}, SparseVLM~\cite{sparsevlm}, VisionZip~\cite{visionzip}, DivPrune~\cite{divprune}, and CDPruner~\cite{cdpruner}. All pruning methods retain 32 visual tokens, and the reported prefill latency includes the complete token-scoring and subset-selection procedure.

\begin{table}[h]
\centering
\caption{\textbf{Detailed efficiency measurements on LLaVA-1.5-7B.} All pruning methods retain 32 visual tokens. Prefill latency includes token-selection overhead, and all quantities are measured under matched inference settings.}
\label{tab:efficiency}
\setlength{\tabcolsep}{3.1pt}
\renewcommand{\arraystretch}{1.05}
\resizebox{\columnwidth}{!}{%
\begin{tabular}{lccccc}
\toprule
Method
& FLOPs \(\downarrow\) (T)
& Prefill \(\downarrow\) (ms)
& Decode \(\downarrow\) (ms/token)
& Resident GPU Mem. \(\downarrow\) (GB)
& MME \(\uparrow\) \\
\midrule
Baseline & 8.94 & 144.6 & 28.0 & 14.5 & 1509.1 \\
FastV & 2.36 & 43.5 & 26.2 & \second{13.9} & 1091.5 \\
PyramidDrop & 3.54 & 68.4 & 25.9 & \second{13.9} & 1204.7 \\
SparseVLM & 2.13 & 48.8 & 31.4 & 18.1 & 1203.2 \\
VisionZip & \best{1.67} & 28.4 & \best{25.1} & 14.0 & 1255.8 \\
DivPrune & \best{1.67} & \second{28.3} & \second{25.3} & \best{13.7} & 1267.9 \\
CDPruner & \second{1.68} & \best{28.2} & \best{25.1} & 15.9 & \second{1362.6} \\
\ourrow \ours & \best{1.67} & \best{28.2} & \best{25.1} & 14.2 & \best{1394.7} \\
\bottomrule
\end{tabular}}
\end{table}

AnchorPrune reduces estimated computation from 8.94 to 1.67~TFLOPs and prefill latency from 144.6 to 28.2~ms, corresponding to an 81.3\% reduction in FLOPs and a \(5.13\times\) prefill speedup relative to the full-token baseline. It also matches the lowest measured decode latency and uses 14.2~GB of resident GPU memory while including the full selection procedure.

CDPruner attains the same measured prefill latency but uses 15.9~GB of resident GPU memory, 1.7~GB more than AnchorPrune and 1.4~GB more than the full-token baseline. Because these values reflect end-to-end resident memory rather than operator-level peak allocation, we do not attribute the difference to a specific algorithmic component. The measurement nevertheless shows that comparable latency and FLOPs do not necessarily imply comparable end-to-end memory behavior.

VisionZip and DivPrune use slightly less resident memory than AnchorPrune, but obtain substantially lower MME scores. PyramidDrop and SparseVLM also remain below AnchorPrune in task performance under the same evaluation setting. CDPruner improves performance over these alternatives but incurs the largest memory footprint among the evaluated pruning methods. AnchorPrune therefore provides the strongest overall operating point in this setting: it matches the lowest measured FLOPs, prefill latency, and decode latency; remains close to the baseline memory footprint; and achieves the highest MME score among all pruning methods.

These results show that AnchorPrune's performance gains do not depend on a computationally or memory-intensive selection mechanism. Its relevance-anchored contextual expansion preserves task performance while maintaining practical end-to-end inference efficiency.

\end{document}